\documentclass[10pt,twocolumn,letterpaper]{article}

\usepackage[pagenumbers]{wacv} % To force page numbers, e.g. for an arXiv version

\usepackage{times}
\usepackage{epsfig}
\usepackage{graphicx}
\usepackage{amsmath}
\usepackage{amssymb}
\usepackage{booktabs}
\usepackage{multirow}
\usepackage{array}
\usepackage{xcolor}
\usepackage{pifont}

\providecommand{\etal}{\emph{et al}.}
\providecommand{\eg}{\emph{e.g}.}

\definecolor{wacvblue}{rgb}{0.21,0.49,0.74}
\usepackage[pagebackref,breaklinks,colorlinks,allcolors=wacvblue]{hyperref}

\usepackage{listings}
\lstdefinestyle{prompt}{
  basicstyle=\ttfamily\scriptsize, breaklines=true, breakindent=0pt,
  columns=fullflexible, keepspaces=true, xleftmargin=4pt, frame=leftline,
  framesep=4pt, rulecolor=\color{gray!50}, aboveskip=4pt, belowskip=4pt,
  literate={—}{{---}}3 {–}{{--}}2 {×}{{$\times$}}1 {’}{{'}}1 {“}{{"}}1 {”}{{"}}1
}

\def\wacvPaperID{1460}
\def\confName{WACV}
\def\confYear{2027}

\begin{document}

%%%%%%%%% TITLE
\title{Autoregressive Mosaics: Probing 2D Spatial Reasoning in Text-Only Language Models}

% Anon. Authors.
%\author{}
\author{Ashwin Nedungadi$^\dagger$ \quad Stefan Oehmcke \quad Stefan L\"udtke \\
Institute for Visual \& Analytic Computing (VAC), University of Rostock\\
{\tt\small \{ashwin.nedungadi, stefan.oehmcke, stefan.luedtke\}@uni-rostock.de}
}
\maketitle
{\let\thefootnote\relax\footnotemark\footnotetext{$^\dagger$ Main author.}}
%%%%%%%%% ABSTRACT
\begin{abstract}
Large language models (LLMs) trained only on text and code can sometimes generate programs that draw recognizable images. However, it is unclear whether this reflects an internal representation of 2D spatial layout or simply the ability to translate spatial descriptions into code. We introduce Autoregressive Mosaics (AM-Bench), a benchmark that separates these factors: First, a translation task gives a model a fully specified geometry of a picture in words as a prompt and asks for the code that produces it. Second, a layout task requires the model to compose an image from an underspecified prompt.  
Across eight open-weight text-and-code-only models, all models reliably translate specified geometry into code, but their open-ended layout performance differs substantially, indicating that these differences are not explained by code-generation ability alone. 
An output-medium ablation further shows that the interface or medium of expression that the model uses matters: replacing procedural code with raw SVG improves layout scores across all models. Finally, probing model activations shows that a coarse layout plan is present before generation, but reflects only the layout implied by the prompt. During generation, models track the evolving geometric state instead of executing an initially fixed plan. 
Overall, these results show that 2D spatial performance in text-only LLMs depends on both the model and the output medium, and is not explained by code-generation ability alone.

\end{abstract}

%%%%%%%%% BODY TEXT
\begin{figure*}[t!]
\begin{center}
\includegraphics[width=0.99\linewidth]{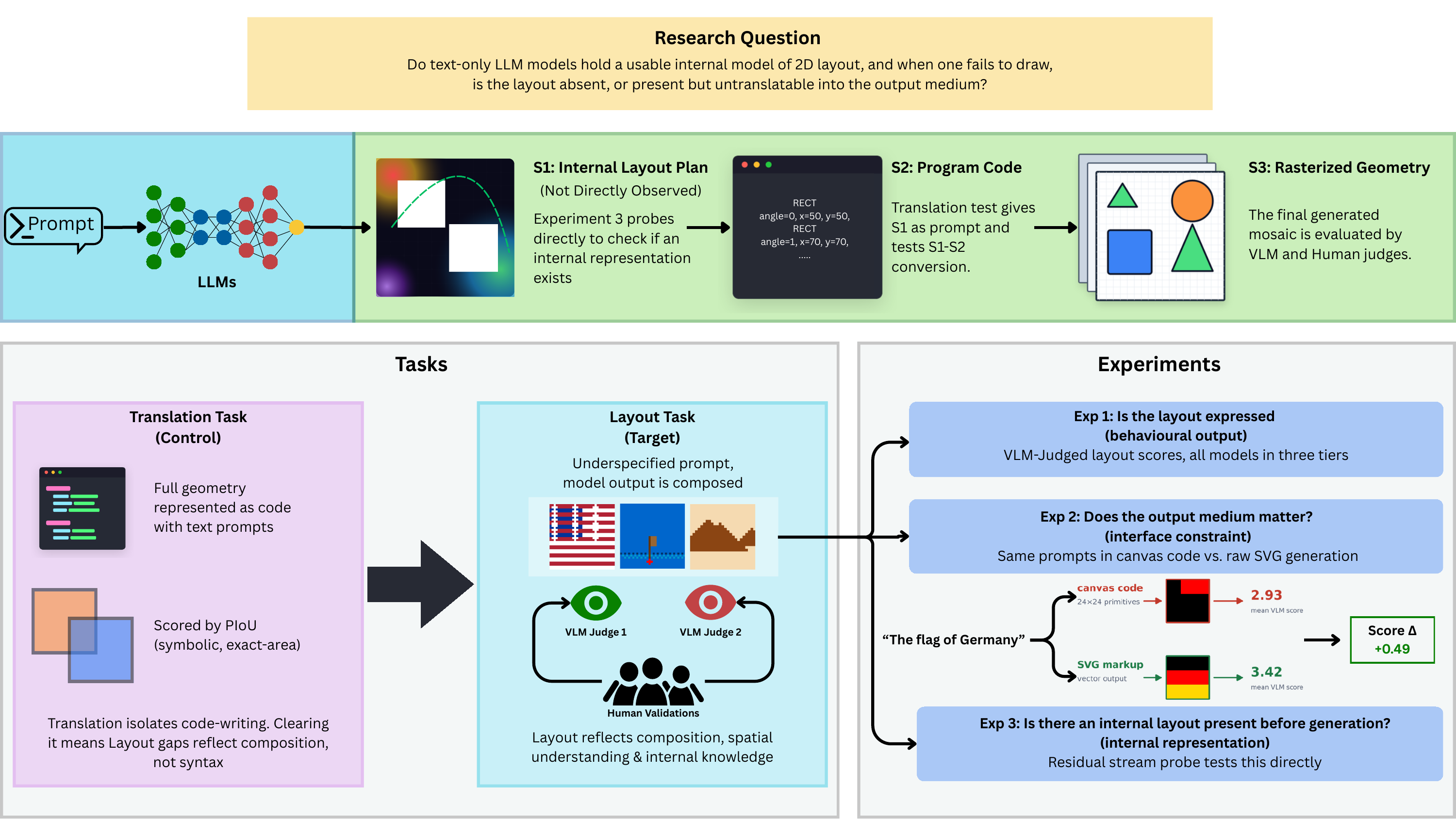}
\end{center}
%\vspace{-3mm}
% NOTE: fig1_overview.pdf graphic to be updated manually by the author to match
% the v5 experiment ordering/numbering (Exp.1 layout, Exp.2 medium, Exp.3 probe).
   \caption{\textbf{Overview of AM-Bench.} The benchmark consists of a four-stage pipeline (prompt, internal plan, program code, rasterized geometry). The translation task states the layout fully in the prompt; in the layout task, layout is underspecified and needs to be decided by the model. Three experiments test expression (Exp.~1), influence of the output medium (Exp.~2), and whether a plan is decodable before generation at all (Exp.~3).}
   %a consistency analysis of repeated generations appears in Appendix~\ref{app:dino_consistency}.}
\label{fig:overview}
\end{figure*}

%------------------------------------------------------------------------
\section{Introduction}
\label{sec:intro}

E\c{s}ref Arma\u{g}an, a painter who is born blind, can draw scenes with linear perspective, occlusion, and consistent shading, with his visual cortex active while he draws~\cite{kennedy2006form,amedi2008blind}. These studies show that in humans, a usable internal model of 2D visual structure exists without ever having seen it. Language models trained only on large scale text and code data exhibit similar abilities: they have never seen an image, yet GPT-4 drew a recognizable unicorn in TikZ when asked \cite{bubeck2023sparks}. This phenomenon has been observed in literature \cite{bubeck2023sparks,li2023othello} but has not been investigated in depth until now. It raises some fundamental questions:
%SL: Okay, but I would not put it in \quote environment, takes a lot of space and seems a bit much
%AN: I agree it takes more space, but it stands out as the central question in the paragraph and draws attention to it. Since the question is interesting and layered, I would want that. I tried giving it more space using \\\\, although essentially I think this is what the quote was doing?

% \emph{To what extent can text-only language models compose and reason about 2D spatial layouts, how does the output medium limit this capability, and does the model construct a useful internal spatial representation, or does it primarily learn to express spatial descriptions through code?}
\begin{quote}
 \emph{To what extent can text-only language models compose and reason about 2D spatial layouts, how does the output medium limit this capability, and does the model construct a useful internal spatial representation, or does it primarily learn to express spatial descriptions through code?}
\end{quote}
An LLM-generated image alone cannot distinguish these possibilities. A model may produce a poor layout because it fails to compose the spatial arrangement, because it cannot express a suitable layout in the required programming language, or because the output medium itself constrains how that layout can be expressed. Existing evaluations typically entangle these factors: recognition benchmarks \cite{dori2026,vrubench2026,ma2024threedsrbench} require visual input, and generative evaluations only assess the final output without separating spatial composition from the ability to translate a spatial description into an executable representation.\\
We therefore introduce Autoregressive Mosaics (AM-Bench, see Fig.~\ref{fig:overview}), a benchmark designed to separate spatial composition from spatial expression. AM-Bench represents an image as an \emph{autoregressive mosaic}, a small raster produced by a program and rendered by a deterministic executor. This evaluates two complementary tasks. In the \emph{translation task}, the prompt specifies the complete geometry of the image, and the model must express that geometry as code. In the \emph{layout task}, the prompt is underspecified and the model must determine the spatial arrangement itself. Thus, translation provides a control for code-generation ability, while layout measures the additional challenge of composing a spatial arrangement. The translation task is scored by an exact-area geometric metric in continuous coordinates, so the score does not depend on canvas resolution. The layout task has no reference geometry to score against, so it is scored by vision language models (VLM), validated by human scorers which provide supporting evidence for the judge's ranking signal. 
Further, we evaluate whether spatial performance is determined by how the model must express its output, by comparing the code interface with raw SVG generation. Finally, we probe model activations before generation to ask whether a coarse spatial layout is represented before any drawing code is produced. We complement this with a causal intervention on the geometric state to test whether models use such a representation during generation.
\\
Across eight open-weight text-and-code-only models, all models reliably solve the translation task, but their performance differs substantially on the open-ended layout tasks, showing that code generation is not the main bottleneck. Second, replacing procedural canvas code with raw SVG improves layout scores across all models, showing that output medium influences results. Third, a coarse spatial representation is present before generation, but it only reflects the layout implied by the prompt. During generation, models instead track the evolving geometric state, consistent with an incremental construction process rather than execution of a fixed layout plan. Overall, these results show that 2D spatial performance in text-only language models depends on both the model and the output medium, and is not explained by code-generation ability alone. We release all code, prompts, and generations on our \textbf{Project Page}\footnote{Hyperlink removed during review process for anonymity}.

%------------------------------------------------------------------------
\section{Related Work}
\label{sec:related}
\noindent\textbf{Emergent structure in text-trained models.} Othello-GPT showed that a transformer trained only on move sequences holds a board representation that can be read out with a probe \cite{li2023othello,nanda2023linear}, later replicated in chess \cite{karvonen2024chess} and revisited with more comprehensive probing across more models \cite{yuan2024revisiting}. A grid that can be recovered from a model that never saw a board is the reason to ask whether a grid-structured \emph{visual} layout exists inside models that never saw an image; our Exp.~3 uses the same probing approach, with capacity-controlled probes and control tasks \cite{hewitt2019control,belinkov2022probing}. Schaeffer \etal\ warns that apparent emergence can be a metric artifact \cite{schaeffer2023mirage}.

\noindent\textbf{Code and structured language as a drawing medium.} Getting neural networks to generate images predates language models. Unlike sketch-rnn~\cite{ha2018sketchrnn}, which trains explicitly on stroke data, the systems studied here write general-purpose code without image-specific training. Following the TikZ unicorn~\cite{bubeck2023sparks}, works like SceneCraft~\cite{hu2024scenecraft}, CoCo~\cite{coco2025}, and LLM Blueprint~\cite{gani2024blueprint} paired LLMs with renderers or refinement loops, adding learned components rather than isolating the model's inherent spatial abilities. More recent work evaluates programmatic visual generation directly: SGP-GenBench scores LLM-written SVG against natural images \cite{sgpgenbench2025}. PRISM finds a large gap between code that runs and code that is spatially correct in programmatic video generation \cite{prism2026}, similar to how we are trying to evaluate both these aspects separately in this work. PlanarBench \cite{planarbench2026}, ASCIIBench \cite{asciibench2025}, DrawingBench \cite{drawingbench2025}, and SVE-ASCII \cite{sveascii2026} test spatial reasoning through ASCII or planar-graph drawing rather than an executable canvas API. Code also appears on the input side of vision: ViperGPT composes vision-language modules by generating Python that is executed against a given image \cite{suris2023vipergpt}. We use code as the output medium instead, to produce a mosaic rather than reason about one. AM-Bench differs from all of these in isolating translation from composition and studying them individually.

\noindent\textbf{Spatial-reasoning benchmarks.} DORI \cite{dori2026}, VRUBench \cite{vrubench2026}, and 3DSRBench \cite{ma2024threedsrbench} require an image as input and are recognition-only. ARC \cite{chollet2019arc} evaluates few-shot learning of programs over grids. T2I-CompBench \cite{huang2023t2icompbench} evaluates diffusion models trained on billions of images. None of them separates translation from composition. Even when frontier models are given an assistive tool for rendering and rotating 3D imagery, spatial-imagery reasoning remains limited \cite{hayashi2026limits}, which is consistent with our results in Exp.~3.

\noindent\textbf{Reference-free text-to-image evaluation.} TIFA \cite{hu2023tifa} and DSG \cite{cho2024dsg} score a generated image against its prompt via question answering, without a reference image, which is similar to  what our layout task needs. However, both still require a real image and a vision-language model that can see it. Additionally, they score whether a picture matches a caption, not whether a picture is geometrically correct, and neither offers a way to isolate composition from code-writing skill the way the translation task does. We use VLM judges for a similar reason (no reference geometry exists for our underspecified prompts), but restrict them to five dimensions and validate them against a translation-verified control task rather than against each other alone.

\noindent\textbf{VLM-as-judge.} MLLM judges approach human agreement on pairwise comparisons but show real biases on absolute scoring \cite{chen2024mllmjudge,zheng2023judging}. We use two independent judges, Qwen2.5-VL-7B \cite{qwen2vl2024} and InternVL3-8B \cite{chen2025internvl3}, to cross-check each other, and restrict each judge to five rubric dimensions rather than to scoring spatial relations directly. Pairwise human-validation for these judges are also presented (Appendix~\ref{app:human}) and will be expanded upon in future work.

\noindent\textbf{SSL representations.} DINO\cite{dinov1, oquab2023dinov2} CLS embeddings support dense prediction with linear heads, and DINO-WM builds a world model capable of planning on frozen features \cite{zhou2024dinowm}. We use DINO embeddings not to score a generation against a reference, but to check how tightly a model's repeated attempts at one prompt cluster together, as a proxy for whether the model is drawing from one stable internal picture. Using it this way, on LLM generated mosaics, is new here (Appendix~\ref{app:dino_consistency}).

%------------------------------------------------------------------------
\section{AM-Bench}
\label{sec:bench}

\subsection{Autoregressive mosaics and the canvas}
\label{sec:executor}
% We define an autoregressive mosaic as one that is produced by an LLM, which is writing a short program that a deterministic executor then renders to a small raster. A custom canvas API built with standard Python libraries consisting of six primitives, \texttt{fill}, \texttt{set\_pixel}, \texttt{rect}, \texttt{circle}, \texttt{line}, \texttt{poly}, is defined over a $24{\times}24$ canvas. The resolution was chosen to bound compute and memory, as the tasks here test spatial composition, where fine rendering precision is not required.
We define an autoregressive mosaic as a short LLM-written program rendered to a $24\times24$ raster via a custom Python API using six primitives (\texttt{fill}, \texttt{set\_pixel}, \texttt{rect}, \texttt{circle}, \texttt{line}, \texttt{poly}). This coarse resolution bounds compute while adequately testing spatial composition.
The model writes one rendering function which is limited to the primitives, arithmetic, iteration, and math. Rasterisation uses Bresenham lines, midpoint-circle fill, and polygon fill. To prevent color tokenization from acting as a confounder, a palette of 30 recommended color names is utilized. General-purpose code generation is itself a well-benchmarked capability \cite{chen2021humaneval} so our translation task (Sec.~\ref{sec:translation_design}) isolates the narrower question of whether that capability transfers to this specific six-primitive vocabulary.

We employ a custom, primitive vocabulary instead of formats like SVG to mitigate data contamination \cite{riddell2024contamination}: Modern LLMs are extensively pretrained on public hand-authored SVG markup \cite{lozhkov2024stackv2} and are increasingly fine-tuned for SVG generation \cite{xing2024llm4svg}. Consequently, evaluating spatial reasoning directly via SVG generation \cite{sgpgenbench2025} risks measuring the retrieval of memorized training patterns rather than true zero-shot layout composition.  We validate this concern in Exp.~2 (Sec.~\ref{sec:exp_medium}), demonstrating that model layout scores artificially inflate when using SVG instead of our API.

\subsection{Translation Task}
\label{sec:translation_design}
The translation task evaluates a model's ability to write rendering code, allowing us to separate it from its ability to spatially compose a mosaic. The prompt states the complete reference geometry in words: every shape, with its position, size, and color, in the same row/column grid the model already uses, using type-and-extent language (\eg\ ``a circle, color red, centered at row~12, column~12, radius~10 cells,'' instead of ``a red sun''). Naming the shape's real-world subject would let a model recall a familiar drawing of that subject instead of reading the geometry it was actually given, so subject names are withheld throughout, and every prompt is checked by hand and by string search to confirm this. A correct response is close to a direct transcription, i.e., nothing is left for interpretation.

\begin{figure}[t]
\begin{center}
\includegraphics[width=1\linewidth]{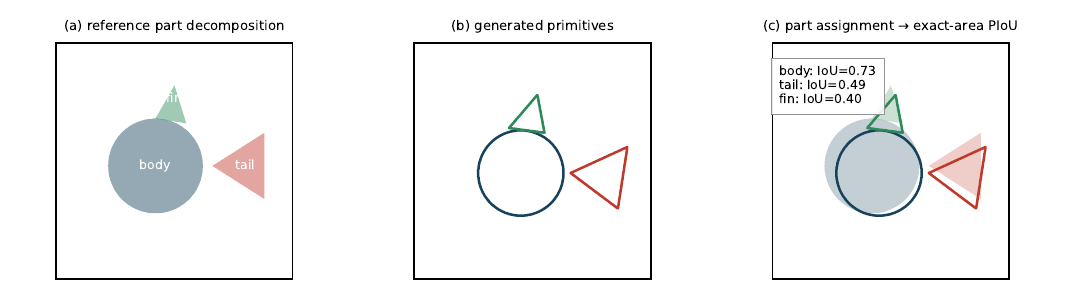}
\end{center}
%\vspace{-3mm}
   \caption{\textbf{Symbolic scoring illustrated.} Generated primitives are assigned to reference parts and scored by exact-area IoU in continuous coordinates. Real translation references are abstract shape configurations, not recognizable subjects (Sec.~\ref{sec:translation_design}); this decomposition illustrates the assignment mechanism, and PIoU is applied only to the translation task, never to Layout.}
\label{fig:scoring}
\end{figure}

Because the reference geometry is known exactly, translation is scored \emph{symbolically}, before rasterization, in normalized $[0,1]^2$ coordinates. Given reference parts $R_1,\dots,R_n$; each generated primitive is assigned to whichever reference part it overlaps most, with ties broken toward the smaller part (Appendix~\ref{app:metric}), and a primitive that does not clear a minimum overlap is left unassigned. Writing the per-part union of assigned primitives as $\hat G_i$,
\begin{equation}
\mathrm{PIoU} = \frac{1}{n}\sum_{i=1}^{n}\frac{|R_i\cap\hat G_i|}{|R_i\cup\hat G_i|}\in[0,1],
\label{eq:piou}
\end{equation}
with areas from exact polygon clipping. The per-part ratio in Eq.~\ref{eq:piou} is the standard Jaccard/intersection-over-union overlap measure~\cite{jaccard1912}, ubiquitous in segmentation evaluation~\cite{everingham2010pascal,long2015fcn}; PIoU is its per-part mean, computed from exact geometry rather than pixel counts. Because no pixel is involved, the score is identical at every resolution (Fig.~\ref{fig:scoring}). An \emph{over-paint} penalty, $\mathrm{PIoU}_{\text{pen}} = \mathrm{PIoU}\times(1 - \text{over-paint fraction})$, checks that a high score is not won by drawing extra, unrequested shapes on top of the correct ones and invalid outputs score zero. See appendix~\ref{app:metric} and~\ref{app:translation_details}.

We built 145 references from a stratified sample of Layout-task generations, then a further 100 built to be harder along four controlled axes plus 13 hand-crafted icons to improve task variety and difficulty (Appendix~\ref{app:translation_details}). The pass threshold $\tau_{\text{trans}} = 0.6$ was chosen before any scores were seen and confirmed well above a chance baseline (Appendix~\ref{app:translation_details}).
%%%%%%%% Fig 2 will be moved to appendix because of lack of space %%%%%%%%
The translation task ensures that low layout scores are not merely artifacts of poor coding ability. However, because this task explicitly hands the model a text-based plan, it only evaluates the mapping of that plan into the output. It cannot decouple an absent internal layout from one that exists but remains unconvertible. To address this directly, Experiment~3 (Sec.~\ref{sec:exp3}) investigates the model's internal activations, shifting the analysis of latent spatial plans from behavioral observation to representation-level probing.

\subsection{Layout Task}
\label{sec:layout_design}
The layout task prompts name an object and at most a coarse appearance, the model must reason and compose the spatial structure itself. No reference exists for such underspecified and open-ended prompts, so PIoU cannot be applied here. The output is instead scored by two VLM judges rating five holistic dimensions (0--5; pass $\geq 3.0$). A model's layout score enters cross-model comparison only if its median translation score clears $\tau_{\text{trans}}$.
% New paragraph describing the VLM judges setup & reasoning
Images are independently evaluated by two VLMs, Qwen2.5-VL-7B~\cite{qwen2vl2024} and InternVL3-8B~\cite{chen2025internvl3}, using an identical system prompt and rubric (Appendix~\ref{app:sp_judge}, \ref{app:sp_rubric}). Each judge outputs discrete scores (0--5) across five axes: prompt fidelity, shape accuracy, color accuracy, spatial accuracy, and completeness (Appendix~\ref{app:judge}). Inter-judge agreement and human validation are detailed in Sec.~\ref{sec:exp1}. We use a VLM-based evaluation over joint embedding distances like CLIPScore~\cite{hessel2021clipscore} or VQAScore~\cite{lin2024vqascore}, as a scalar similarity metric inherently conflates color, shape, and spatial errors. Furthermore, standard dual encoders exhibit bag-of-words behavior regarding compositional and spatial relations~\cite{yuksekgonul2023bagofwords}, which is the failure mode we attempt to isolate.

%% Table doesn't add much and takes up space %%
% \begin{table}[h]
% \centering
% \small
% \setlength{\tabcolsep}{5pt}
% \begin{tabular}{@{}lll@{}}
% \toprule
%  & High layout & Low layout \\
% \midrule
% High translation & layout present & \textbf{layout absent, or}\\
%  & & \textbf{not convertible} \\
% Low translation & (n/a, gated out) & translation-limited \\
% \bottomrule
% \end{tabular}
% \caption{Only the high-translation row supports a claim about internal layout.}
% \label{tab:interp}
% \end{table}

% Reviewer Q: Is this better presented as a table?
\subsection{Prompt Suite}
\label{sec:suite}
The layout task has 150 prompts per model categorized into three tiers (10 prompts for 5 subcategories):

\noindent\textbf{T1 Elemental} (1A--1E): single geometric primitives or motifs: filled shapes (1A), hollow outlines (1B), patterns and tilings (1C), compound symbols (1D), color partitions (1E).

\noindent\textbf{T2 Iconic} (2A--2E): recognisable objects: flags (2A), cultural symbols (2B), everyday objects (2C), nature and living things (2D), structural patterns (2E).

\noindent\textbf{T3 Compositional} (3A--3E): multi-element compositions: binary spatial relations (3A), multi-element arrangements (3B), nested or hierarchical structures (3C), scene compositions (3D), symmetry and transformation (3E).

\noindent\textbf{Prompt construction.} The categories were hand-crafted and all 150 prompts were synthetically generated (using Claude Opus 5), motivated by the number of required prompts: Layout alone requires 150 prompts $\times$ 8 models $\times$ 11 samples (13,200 generations), while keeping wording and difficulty consistent within a subcategory. A small held-out set of human-written prompts, produced early in the project, seeded the process as worked examples of each tier's intended phrasing and difficulty, but were kept out of the released 150-prompt suite (See Fig.~\ref{fig:human_mosaics}, Fig.~\ref{fig:synthetic_mosaics}).

%------------------------------------------------------------------------
\section{Experiments}
\label{sec:experiments}
We evaluated eight open-weight, text- and code-only models spanning four families and varying parameter counts (8B--34B): Qwen2.5-Coder 32B/14B, Gemma~2 27B/9B, GLM-4 32B/9B, CodeLlama~34B, and Llama~3.1 8B, one model loaded at a time. Layout samples 11 generations per prompt (1 deterministic, 10 stochastic; $N_{\text{stochastic}}{=}10$), for 150 prompts $\times$ 8 models $\times$ 11 samples $=$ 13,200 attempts. Of these, \textbf{12,932 (98.0\%) are valid} (i.e., the generated code executes and renders without error) and 268 (2.0\%) fail to execute or render. Within the valid set, \textbf{11,976 (92.6\%) are non-trivial} and 956 (7.4\%) are trivial, meaning the code runs but produces a degenerate near-uniform fill where over 98\% of the $24{\times}24$ pixels have the same color. Both VLM judges score every valid attempt, trivial ones included (12,932 for Judge~1; 12,926 for Judge~2, six images Judge~2 could not process, spread across five models).  
Translation produces 13,920 attempts on the 145 reference set (145 references $\times$ 8 models $\times$ 12 samples). 
\subsection{Exp.~1: Is the layout there?}
\label{sec:exp1}

Since all models passed the translation task (Table~\ref{tab:validity}), which controls for code-writing competence, the variance in the subsequent layout scores isolates differences in spatial composition rather than code-generation capability.

\noindent\textbf{Validity and trivial rates.}
Every model has valid output across all tiers (Table~\ref{tab:validity}; worst case 88.2\%), so code is a viable medium with no parsing bottleneck for any of the models. Trivial (uniform fill) rates vary far more than validity: CodeLlama~34B produces trivial outputs on 12--27\% of its attempts, the highest of any model at T1 and T2, while Gemma 2 9B spikes to 20.5\% at T3, the single highest trivial rate in the table. These suggest a fallback to a uniform fill when the model is least confident about the geometry it should compose.

\begin{table}[t]
\centering
\footnotesize
\setlength{\tabcolsep}{2.8pt}
\renewcommand{\arraystretch}{0.95}
\begin{tabular}{@{}lccc@{\hspace{5pt}}ccc@{\hspace{5pt}}c@{}}
\toprule
& \multicolumn{3}{c}{Valid\%} & \multicolumn{3}{c}{Trivial\%} & Transl. \\
\cmidrule(r){2-4}\cmidrule(lr){5-7}\cmidrule(l){8-8}
Model & T1 & T2 & T3 & T1 & T2 & T3 & PIoU \\
\midrule
GLM-4 32B          & \phantom{0}94.5 & \phantom{0}94.4 & \phantom{0}99.5  & \phantom{0}1.3 & \phantom{0}1.1 & \phantom{0}1.6 & 0.993 \\
Qwen2.5-Coder 32B  & 100.0 & 100.0 & \phantom{0}99.8 & \phantom{0}5.8 & \phantom{0}0.9 & \phantom{0}4.4 & 0.994 \\
Qwen2.5-Coder 14B  & 100.0 & 100.0 & 100.0 & \phantom{0}6.4 & \phantom{0}2.2 & \phantom{0}3.5 & 0.995 \\
Gemma~2 27B        & 100.0 & \phantom{0}99.5 & 100.0 & \phantom{0}9.6 & \phantom{0}0.5 & \phantom{0}4.4 & 0.992 \\
GLM-4 9B           & \phantom{0}99.1 & \phantom{0}93.5 & 100.0 & \phantom{0}6.7 & \phantom{0}1.5 & \phantom{0}5.1 & 0.991 \\
Gemma~2 9B         & \phantom{0}93.5 & \phantom{0}95.3 & \phantom{0}88.2 & 13.8 & \phantom{0}3.3 & 20.5 & 0.980 \\
Llama~3.1 8B       & \phantom{0}99.8 & \phantom{0}99.5 & 100.0 & \phantom{0}9.8 & \phantom{0}4.2 & 10.2 & 0.929 \\
CodeLlama~34B      & \phantom{0}98.7 & \phantom{0}98.2 & \phantom{0}98.0 & 26.9 & 12.4 & 17.8 & 0.984 \\
\bottomrule
\end{tabular}
\caption{\textbf{Layout validity, trivial rates, and translation PIoU.} \textbf{Validity \& Trivial Rates:} All models exceed 88\% validity (successful execution and rendering) across all tiers ($n{=}550$/model/tier). Outputs that render a uniform color (trivial) provide a stronger failure signal, peaking in CodeLlama~34B and Gemma~2 9B. \textbf{Transl.\ PIoU:} Mean per-part IoU on the translation set ($n{=}1{,}740$/model; Sec.~\ref{sec:translation_design}), independent of tier.  The \emph{median} PIoU is 1.000 for all models, passing the threshold ($\tau_{\text{trans}}{=}0.6$). Overall, 98.1\% of all 13,920 attempts pass, far exceeding the 0.072 random baseline.}
\label{tab:validity}
\end{table}

\noindent\textbf{VLM layout scores.}
Table~\ref{tab:scores} shows mean VLM scores under both judges. GLM-4~32B leads at every tier under both judges (J1 overall 3.57), down to CodeLlama~34B, last at every tier (J1 overall 1.75), identical under J1 and J2. Judge~2 (InternVL3-8B) is systematically about a point more lenient than Judge~1 (Qwen2.5-VL-7B), but the two judges agree exactly on model-level rank ordering (Spearman $\rho{=}1.0$) despite only moderate per-sample agreement (inter-judge Pearson $r{=}0.43$--$0.54$, Spearman $\rho{=}0.42$--$0.57$ across the eight models). The benchmark's model ranking is robust even though individual sample scores carry real judge-to-judge noise. Over a fixed 350 pair set, two independent annotators each judged all 350 pairs (Krippendorff's $\alpha = 0.60$)\cite{krippendorff2004content}, agreeing on 70.6\% of pairs. On pairs where both annotators agree and the judge separates the pair, Judge1 matches the human verdict on 82.4\% (61/74) and Judge2 on 76.7\% (46/60) (Appendix~\ref{app:human}).
% \begin{table}[t]
% \centering
% \footnotesize
% \setlength{\tabcolsep}{3pt}
% \renewcommand{\arraystretch}{0.9}
% \begin{tabular}{@{}lcccc@{}}
% \toprule
% Model & T1 & T2 & T3 & Overall \\
% \midrule
% \multicolumn{5}{@{}l}{\textit{J1 (Qwen2.5-VL-7B)}} \\
% GLM-4 32B         & 3.93 & 3.14 & 3.64 & 3.57 \\
% Qwen2.5-Coder 32B & 3.79 & 3.07 & 3.46 & 3.44 \\
% Qwen2.5-Coder 14B & 3.66 & 3.02 & 3.22 & 3.30 \\
% Gemma~2 27B       & 3.31 & 2.28 & 3.18 & 2.92 \\
% GLM-4 9B          & 3.27 & 2.59 & 2.76 & 2.87 \\
% Gemma~2 9B        & 3.13 & 2.09 & 2.00 & 2.41 \\
% Llama~3.1 8B      & 2.41 & 1.91 & 1.79 & 2.03 \\
% CodeLlama~34B     & 2.31 & 1.48 & 1.47 & 1.75 \\
% \midrule
% \multicolumn{5}{@{}l}{\textit{J2 (InternVL3-8B)}} \\
% GLM-4 32B         & 4.55 & 4.13 & 4.31 & 4.33 \\
% Qwen2.5-Coder 32B & 4.53 & 3.96 & 4.03 & 4.17 \\
% Qwen2.5-Coder 14B & 4.35 & 3.93 & 4.11 & 4.13 \\
% Gemma~2 27B       & 4.15 & 3.43 & 3.91 & 3.83 \\
% GLM-4 9B          & 4.13 & 3.44 & 3.65 & 3.74 \\
% Gemma~2 9B        & 4.09 & 3.18 & 2.70 & 3.32 \\
% Llama~3.1 8B      & 3.65 & 2.81 & 2.90 & 3.12 \\
% CodeLlama~34B     & 3.54 & 2.55 & 2.53 & 2.87 \\
% \bottomrule
% \end{tabular}
% \caption{\textbf{Mean VLM layout scores (0--5) by model $\times$ tier, all eight models.} Pass threshold $\geq 3.0$. J2 is more lenient; both judges agree exactly on model-level rank ordering.}
% \label{tab:scores}
% \end{table}

\begin{table}[t]
\centering
\footnotesize
\setlength{\tabcolsep}{3pt}
\renewcommand{\arraystretch}{0.9}
\begin{tabular}{@{}lcccc@{}}
\toprule
Model & T1 & T2 & T3 & Overall \\
\midrule
\multicolumn{5}{@{}l}{\textit{J1 (Qwen2.5-VL-7B)}} \\
GLM-4 32B         & 3.93 & 3.14 & 3.64 & 3.57 \\
Qwen2.5-Coder 32B & 3.79 & 3.07 & 3.46 & 3.44 \\
Qwen2.5-Coder 14B & 3.66 & 3.02 & 3.22 & 3.30 \\
Gemma~2 27B       & 3.31 & 2.28 & 3.18 & 2.92 \\
GLM-4 9B          & 3.27 & 2.59 & 2.76 & 2.87 \\
Gemma~2 9B        & 3.13 & 2.09 & 2.00 & 2.41 \\
Llama~3.1 8B      & 2.41 & 1.91 & 1.79 & 2.03 \\
CodeLlama~34B     & 2.31 & 1.48 & 1.47 & 1.75 \\
\cmidrule(lr){1-5}
\textbf{Pooled}   & 3.23 & 2.45 & 2.69 & 2.79 \\
\midrule
\multicolumn{5}{@{}l}{\textit{J2 (InternVL3-8B)}} \\
GLM-4 32B         & 4.55 & 4.13 & 4.31 & 4.33 \\
Qwen2.5-Coder 32B & 4.53 & 3.96 & 4.03 & 4.17 \\
Qwen2.5-Coder 14B & 4.35 & 3.93 & 4.11 & 4.13 \\
Gemma~2 27B       & 4.15 & 3.43 & 3.91 & 3.83 \\
GLM-4 9B          & 4.13 & 3.44 & 3.65 & 3.74 \\
Gemma~2 9B        & 4.09 & 3.18 & 2.70 & 3.32 \\
Llama~3.1 8B      & 3.65 & 2.81 & 2.90 & 3.12 \\
CodeLlama~34B     & 3.54 & 2.55 & 2.53 & 2.87 \\
\cmidrule(lr){1-5}
\textbf{Pooled}   & 4.12 & 3.43 & 3.52 & 3.69 \\
\bottomrule
\end{tabular}
\caption{\textbf{Mean VLM layout scores (0--5) by model $\times$ tier, all eight models.} Pass threshold $\geq 3.0$. Every value is the macro-average over the 15 subcategory means; \emph{Pooled} averages that over the eight models, and is the source of the tier means quoted in Sec.~\ref{sec:exp1}. J2 is more lenient by 0.90 overall, but both judges agree exactly on model-level rank ordering, and both show the T2/T3 inversion in the pooled row.}
\label{tab:scores}
\end{table}

% \paragraph{The T2/T3 inversion.} The expected T1$>$T2$>$T3 difficulty gradient does not hold: under J1, mean T3 Compositional (2.69) outscores mean T2 Iconic (2.45), averaged across all eight models. This inversion holds within five of the eight models individually (largest at Gemma~2 27B, $-$0.90; smallest reversal at Gemma~2 9B, CodeLlama~34B, and Llama~3.1 8B, which keep the expected T2$>$T3 order by a small margin). Compositional spatial language is more tractable than iconic object rendering at $24{\times}24$ resolution: recognizable real-world objects need fine structure that geometric code primitives approximate poorly. Note that the \emph{cross-model} spread is actually widest at T3 (range 2.17, vs.\ 1.66 at T2 and 1.62 at T1) even though T2 is the lower-scoring tier on average --- models disagree with each other most on Compositional scenes, while converging more tightly, at a lower level, on Iconic ones. The per-subcategory breakdown (Fig.~\ref{fig:radar_overlay}) agrees on tier difficulty: the hardest subcategory is 2C Everyday Objects (mean 1.97 across all eight models), then 3D Scene Compositions (2.18); the easiest is 1E Color Partitions (3.93).
% Fixed New

\begin{figure*}[hbt]
\begin{center}
\includegraphics[width=0.95\linewidth]{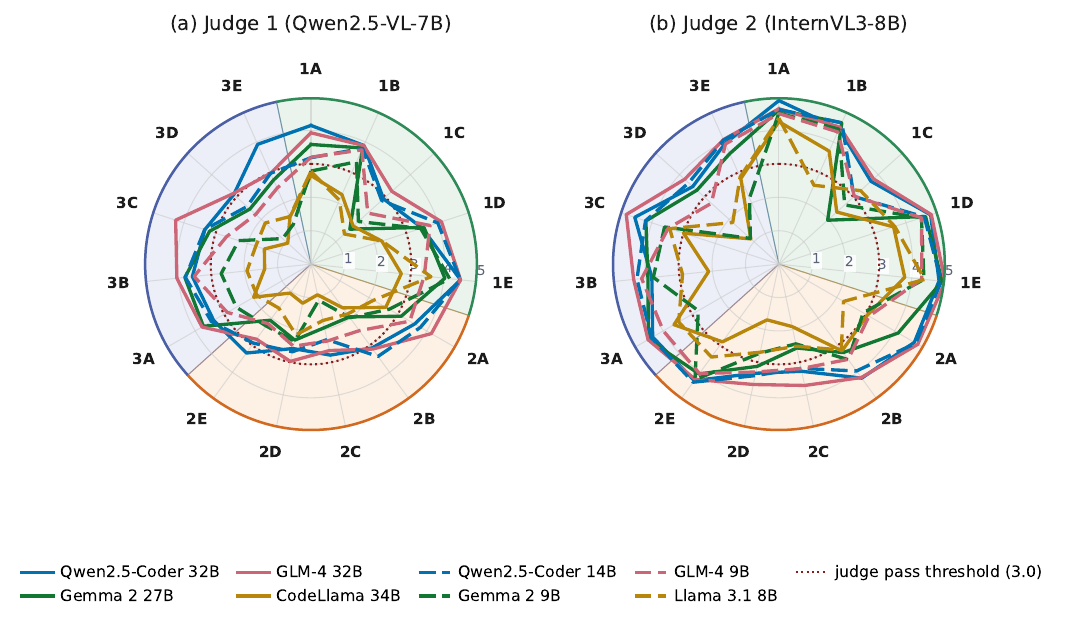}
\end{center}
%\vspace{-3mm}
   \caption{\textbf{Per-subcategory VLM scores, both judges (15 subcategories $\times$ 8 models).} (a) Judge 1 (Qwen2.5-VL-7B); (b) Judge 2 (InternVL3-8B). 2C Everyday Objects and 3D Scene Compositions are the two hardest subcategories and 1E Color Partitions the easiest under both judges. Background color separates tier categories. See Appendix~\ref{app:judge} for detailed plots.}
\label{fig:radar_overlay}
\end{figure*}

\noindent\textbf{The T2/T3 inversion.} The expected T1$>$T2$>$T3 difficulty gradient fails under both judges: mean T3 Compositional outscores T2 Iconic across all models (J1: 2.69 vs.\ 2.45; J2: 3.52 vs.\ 3.43). Defining $\Delta = \text{T2} - \text{T3}$, this inversion ($\Delta < 0$) holds for five models under J1 and six under J2. Seven models exhibit cross-judge directional agreement: Gemma~2 27B inverts most strongly (J1: $-$0.90, J2: $-$0.48), Gemma~2 9B and CodeLlama~34B maintain the expected order, and Llama~3.1 8B splits (J1: $+$0.12, J2: $-$0.09). We attribute this to resolution constraints: recognizing real-world iconic objects requires fine structure poorly approximated by $24{\times}24$ code primitives, making compositional spatial language more tractable. Furthermore, cross-model spread peaks at T3 under both judges (J1 ranges: 2.17 [T3] vs.\ 1.66 [T2], 1.62 [T1]; J2: 1.78 vs.\ 1.58, 1.01). Models thus diverge most on compositional scenes while converging tightly on poor iconic rendering. Subcategory analysis (Fig.~\ref{fig:radar_overlay}) confirms this: 1E Color Partitions is the easiest (J1: 3.93), while 2C Everyday Objects and 3D Scene Compositions are consistently the hardest, though their exact bottom-two ranking swaps between J1 (1.97, 2.18) and J2 (2.85, 2.63).

Fig.~\ref{fig:passatk}): the three strongest models (GLM-4~32B, Qwen2.5-Coder~32B/14B) reach 0.92--0.93 overall and 0.96 at T3, while the weakest (Gemma~2~ 9B, CodeLlama~34B) plateau at 0.66 and 0.69, so $k{=}10$ does not saturate the benchmark (See Appendix~\ref{app:judge}).
Finally, while VLM scores capture adherence to the prompt, we also evaluate whether models draw from a stable internal spatial representation across repeated attempts; this representation-level consistency analysis (using DINO embeddings) is detailed in Appendix~\ref{app:dino_consistency} due to strong confounding effects from token-generation entropy.

\noindent\textbf{Failure analysis.}
Color accuracy is the highest-scoring dimension for every model under both judges. The other four dimensions (prompt fidelity, shape accuracy, spatial accuracy, completeness) track each other closely per model, with no single dimension consistently weakest across the models (Appendix~\ref{app:judge}).

\subsection{Exp.~2: Does the output medium matter?}
\label{sec:exp_medium}
Exp.~1 reads layout scores as evidence about spatial composition, holding the medium (procedural canvas code) fixed. That leaves open whether the medium is neutral or an active constraint on what a model can express. We ablate it for the full 150-prompt Layout suite (all 15 subcategories, all 3 tiers): every model draws the same prompt twice under an otherwise identical protocol (same judge, same rubric, same 1 deterministic $+$ 10 stochastic sampling schedule), once writing canvas-primitive code and once writing raw SVG generation. The canvas baseline is re-scored under the same judge prompt as the SVG arm, with the medium-specific framing removed from the judge's system message (Appendix~\ref{app:system_prompts}).
Direct comparisons between low-resolution Canvas mosaics ($24{\times}24$, nearest-neighbor) and high-resolution, anti-aliased SVGs conflate spatial expressiveness with rendering fidelity. To only measure expressiveness, we eliminate this rendering discrepancy by re-rasterizing all SVGs at $24{\times}24$ from their vector sources before scoring.

\begin{figure}[hbt]
\begin{center}
\includegraphics[width=1.0\linewidth]{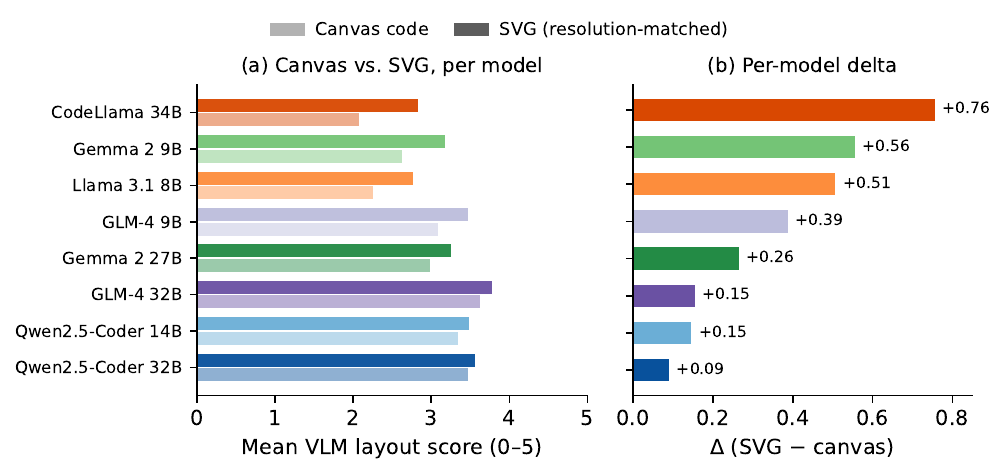}
\end{center}
%\vspace{-3mm}
   \caption{\textbf{Medium ablation, all eight models.} (a)~Canvas vs.\ SVG mean score. (b)~Per-model delta, sorted; every model improves under SVG.}
\label{fig:svg}
\end{figure}

\noindent\textbf{Result.} Fig.~\ref{fig:svg} and Table~\ref{tab:svg_corrected} (Appendix~\ref{app:deep_dive}) show the results. Pooled across all eight models, SVG scores higher: $+$0.37 (95\% CI $[0.26, 0.47]$). Six of eight models are individually significant. This is not a code-model effect, in general, as CodeLlama~34B shows the largest improvement. We hypothesize that the data exposure argument described in Sec.~\ref{sec:executor} explains this result: CodeLlama~34B's training data plausibly include more SVG data~\cite{lozhkov2024stackv2,xing2024llm4svg} (but cannot include our rendering primitives). 
%An independent correction, point-sampling instead of re-rendering, agrees ($+$0.38, CI $[0.17, 0.63]$) and differs from the primary estimate only in the judge's \emph{color} dimension, consistent with an anti-aliasing artifact rather than a correction-method artifact.

This experiment shows that the results depend on the rendering medium. This does not overturn Exp.~1's ranking (the same eight models keep a similar relative order under either medium), but it does mean the absolute layout scores reported anywhere in this paper are a property of the canvas-code interface, not of spatial reasoning in general. 
%
%
% Either keep this or the table.
\subsection{Exp.~3: Is a plan present before generation?}
\label{sec:exp3}
The translation task shows that all models can convert a stated plan into primitives (Sec.~\ref{sec:exp1}), and Exp.~2 shows that the medium restricts what reaches the image. Neither establishes whether a layout plan exists \emph{before} the model writes any code. We test this with a linear probe: a single forward pass over each of the 145 Layout prompts\footnote{The prompts are inherited from the Translation reference set.} gives us the residual-stream activations at every layer. We then train ridge regression on these activations to predict a coarse $6{\times}6$ occupancy grid over the $24{\times}24$ canvas\footnote{To balance target dimensionality and sample size (36 cells need to be estimated from 145 prompts). It also divides the $24{\times}24$ canvas evenly into $4{\times}4$-pixel cells.} \cite{belinkov2022probing}. Critically, rather than a discrete binary label, each cell is treated as a continuous probability map representing the empirical fraction of a model's repeated attempts that draw into it; this continuous $[0,1]$ target dictates the use of regression and $R^2$ variance over classification accuracy. As a prompt alone can already imply where some shape will be drawn (``a house on a hill''), decodability on its own is not informative: we report $\Delta R^2$, the gap over a text-only baseline (the same regression fit on the prompt's TF-IDF encoding instead of on activations). The full method is described in Appendix~\ref{app:exp3_appendix}.
The necessity of this continuous target arises because the binary occupancy of a single reference generation lacks a well-defined ground truth: an underspecified prompt admits many valid layouts, making one discrete generation merely a single sample from a broader distribution. 
%, and all 145 references originate from two models, leaving seven of the eight roster models probed against another model's output. 
A model's own continuous consensus (the average occupancy over all of that model's valid generations for a prompt) predicts the single reference at 0.688 on average, against a text baseline of 0.625. In four of the eight models, that difference is less than 0.05. We therefore use each model's own consensus occupancy as the target instead. Layer, token position, and ridge penalty are selected inside training folds only, and the text baseline is refit per fold, so neither arm is tuned on test data \cite{hewitt2019control}. Against this target, decodability exceeds the text baseline for all eight models after Benjamini-Hochberg correction (mean $\Delta R^2 = {+}0.164$).

\noindent\textbf{What is encoded is generic, not model-specific.} A probe that predicts the generated layout may only be reading what the prompt implies, i.e., layout independently of the model. We separate model-specific and shared (model-agnostic) components: The \emph{shared} component is the leave-one-model-out consensus of the other seven models, and the \emph{specific} component is the residual after subtracting the shared component. Table~\ref{tab:decomposition} reports both for the two strongest large models. The shared component decodes far above text ($\Delta R^2 = {+}0.26$ and ${+}0.28$); the specific component does not ($R^2 \leq 0.005$). Split-half reliability (splitting each model's generations in half and checking how well one half predicts the other) bounds what a probe could recover from this target at 0.883 and 0.941, which are both far above the observed $R^2$. Furthermore, prompt length does not predict the target ($R^2 = -0.03$), and the TF-IDF baseline agrees with a mean token embedding baseline (0.135 TF-IDF, 0.132 mean token embeddings).

\begin{table}[hbt!]
\centering\footnotesize
\setlength{\tabcolsep}{4pt}
\begin{tabular}{@{}llccc@{}}
\toprule
Model & Component & Prompt & Residual & Ceiling \\
\midrule
\multirow{3}{*}{GLM-4 32B}
 & own      & 0.121 & 0.318$^{***}$ & 0.955 \\
 & shared   & 0.289 & \textbf{0.550}$^{***}$ & --- \\
 & specific & $-$0.014 & \textbf{0.005}\phantom{$^{***}$} & \textbf{0.883} \\
\midrule
\multirow{3}{*}{Gemma 2 27B}
 & own      & 0.085 & 0.306$^{***}$ & 0.978 \\
 & shared   & 0.293 & \textbf{0.573}$^{***}$ & --- \\
 & specific & $-$0.029 & \textbf{$-$0.005}\phantom{$^{***}$} & \textbf{0.941} \\
\bottomrule
\end{tabular}
% \caption{\textbf{Decomposed decodability of the pre-generation residual stream.} Out-of-fold pooled $R^2$ under nested cross-validation. \emph{Text} is a train-fold-only TF-IDF baseline, \emph{Ceiling} the split-half measurement ceiling. $^{***}$: $p<0.001$, prompt-clustered bootstrap with BH-FDR. The shared component is decoded far above text, the model-specific component is not.}
%New
\caption{\textbf{Pre-generation residual stream decodability.} Nested cross-validated pooled $R^2$. We decompose a model's \emph{own} consensus layout into a \emph{shared} component (consensus of the other seven models) and a \emph{specific} remainder (\emph{own} $-$ \emph{shared}). \emph{Prompt} is a TF-IDF baseline; \emph{Ceiling} is the split-half reliability bound. $^{***}$: $p<0.001$ (prompt-clustered bootstrap, BH-FDR). Crucially, \emph{own} decodability is driven entirely by the \emph{shared} prompt layout: the model-\emph{specific} realization is undecodable, despite the \emph{Ceiling} confirming the target is measurable.}
\label{tab:decomposition}
\end{table}
%\vspace{-0.5cm}

\noindent\textbf{Does the model use what it encodes?} Decodability shows a signal is present, not that the model relies on it \cite{li2023othello}, so we test use directly with an intervention on GLM-4~32B and Gemma-2~27B. For each of the model's own generations, we truncate the code after its first drawing statement and compare four conditions resumed from this identical prefix: \textbf{control} appends nothing, \textbf{null\_asis} appends the model's own real next statement, \textbf{moved\_unused} appends that statement with its textual coordinates edited so the shape lands where the original generation does not draw, and \textbf{moved\_used} moves it instead onto a region the original generation was going to fill later. 731 generations across the two models meet our eligibility bar (enough spatial room for both edits). Each one is resumed 9 times per arm (8 stochastic, 1 greedy), giving 2{,}924 (item, arm) pairs and 26{,}316 continuations. A model executing a plan formed before it started drawing should treat a displaced shape as an error to correct, while a model without a pre-formed plan should treat it the same as the original.
%
% Concise version
For completed generations, the model typically stops writing code when the textual geometric state remains unmodified (\texttt{null\_asis}). We evaluate whether displacing a completed shape into an unused region (\texttt{moved\_unused}) prompts the model to resume generating drawing primitives. This textual coordinate perturbation significantly increases continuation rates: GLM-4~32B rises from 43.8\% to 84.6\% ($\Delta = {+}0.42$), and Gemma-2~27B from 2.5\% to 41.2\% ($\Delta = {+}0.39$, both BH-FDR $p = 0.0007$). The effect scales with displacement magnitude: a smaller shift into a region the model intended to fill later (\texttt{moved\_used}, median 4\,px vs.\ 9\,px for \texttt{moved\_unused}) yields a weaker continuation response (${+}0.08$ for GLM-4~32B; ${+}0.16$ for Gemma-2~27B).
Because the perturbations causing this effect are strictly textual digit substitutions within the code sequence, this behavior demonstrates that the models actively track the textually encoded spatial completeness of the programmatic prefix, rather than pattern-matching raw code syntax. Furthermore, this behavior shows that layout composition is an incremental, autoregressive process governed by the textual geometric state, not solely the execution of an \emph{a priori} spatial plan. 
%\vspace{-0.5cm}

\noindent\textbf{Scope.} The shared/specific decomposition (Table~\ref{tab:decomposition}) covers only two of the eight models, while the eight model decodability result uses the undecomposed target only. Furthermore, the causal test intervenes on the textual geometric state, not on internal activations. It shows generation is sensitive to what has already been encoded in the code prefix, not that the decoded representation drives that sensitivity, which would require intervening on activations directly (e.g.\ activation patching).

\section{Conclusion}
\label{sec:conclusion}
\label{sec:limitations}
AM-Bench separates the two quantities that a single pass/fail score mixes together: whether a usable internal layout exists, and whether the model can express it through the medium it needs to use.

\noindent\textbf{Translation is not the bottleneck.} The translation task shows that all models can express layouts in the medium: median PIoU is 1.0 for all models. The sharp layout-score differences that follow (where GLM-4~32B generally performed best and Compositional prompts are easier than Iconic ones in five of the eight models) are therefore not explained by an inability to write correct code. Those layout scores rest on two VLM judges, which two annotators matched on 82.4\% and 76.7\% of the pairs where the annotators agreed with each other (Appendix~\ref{app:human}). While the absolute scores carry judge-specific differences, they produced the same ranking.

% \noindent\textbf{The medium affects the performance.} Swapping our custom canvas medium for raw SVG raises every model's score, even when controlling for SVG's higher native rendering fidelity. Because modern LLMs are extensively pretrained on public SVG markup, this score improvement is heavily confounded by pretraining data memorization. This contrast validates our core design choice: evaluating true zero-shot spatial composition requires a custom, uncontaminated interface like AM-Bench's canvas API, as standard formats conflate emergent spatial reasoning with data retrieval. 
\noindent\textbf{The medium affects performance.} Swapping our custom canvas medium for raw SVG raises every model's score, even when controlling for SVG's higher native rendering fidelity. Because modern LLMs are extensively pretrained on public SVG markup, prior exposure to the format introduces a potential confound. This motivates our use of a custom, uncontaminated interface like the AM-Bench canvas API to evaluate zero-shot spatial composition, ensuring we measure emergent spatial reasoning rather than potential data retrieval.

\noindent\textbf{Only a generic plan is encoded.} We find that a coarse layout is linearly decodable above a text baseline for all eight models. To determine what this decoded signal actually represents, we decompose it, for the two strongest models, into a component shared with other models' outputs for the same prompt and a component specific to the probed model. Only the shared component is decodable since the model-specific component reaches $R^2 \leq 0.005$. A behavioral causal test on the same two models further shows that, during generation, models actively track the textually encoded geometric state instead of executing a spatial plan fixed in advance.
\\\\
\noindent\textbf{Limitations.} In this paper, we evaluate eight open-weight models (8B--34B), but generalization to larger or closed-weight models remains untested. AM-Bench assesses spatial competence by evaluating whether the models can generate code that draws small mosaics. While this ability can be seen as a necessary condition, it does not test spatial competence more generally. 
Furthermore, the VLM-as-judge evaluation introduces variance (although the judges agree on the ranking of the methods and are verified by human validation). Finally, Exp.~3 intervenes on the generated code, but not on internal activations, which would allow to investigate internal spatial planning more directly.\\\\
%(See Appendix~\ref{app:limitations_full}).
%
In summary, the programmatic generation of 2D spatial layouts depends on both the model's compositional ability and the output medium. Before generation, the model encodes the generic layout its prompt implies, but not the particular layout it will draw, which is instead decided autoregressively as generation proceeds. The same design could be extended to 3D, e.g.\ through structured-language media~\cite{avetisyan2024scenescript}.
{
\small % according to official template
\bibliographystyle{ieee_fullname}
\bibliography{ambench}
}

\clearpage
\appendix
\twocolumn[{\begin{center}{\large\bf Supplementary Material}\end{center}}]

\section{Model Generation \& Prompting}
\label{app:generation_prompting}

\subsection{Medium specifications}
\label{app:mediums}
The system prompt defines the six primitive signatures, the \texttt{canvas.rows}/\texttt{cols} attributes, the sandbox restrictions (arithmetic, iteration, \texttt{math}; no imports or I/O), the named 30-color palette, and the single-\texttt{render(canvas)} requirement. AST violations, runtime exceptions, or timeouts render an error pattern (validity logged).

\subsection{System prompts}
\label{app:system_prompts}

All prompts are reported verbatim. Table~\ref{tab:prompt_matrix} maps each experiment to the prompts it used; only two generation prompts and two judge system prompts exist across the paper.

\begin{table}[h]
\centering\footnotesize
\setlength{\tabcolsep}{3pt}
\begin{tabular}{@{}lll@{}}
\toprule
Experiment & Generation prompt & Judge system prompt \\
\midrule
Translation  & Canvas (\ref{app:sp_canvas})$^{\dagger}$ & --- (symbolic) \\
Exp.~1 Layout    & Canvas (\ref{app:sp_canvas}) & Pixel-art (\ref{app:sp_judge}) \\
Exp.~2 canvas arm& Canvas (\ref{app:sp_canvas}) & Generic (\ref{app:sp_generic}) \\
Exp.~2 SVG arm   & SVG (\ref{app:sp_svg})       & Generic (\ref{app:sp_generic}) \\
Exp.~3  & Canvas (\ref{app:sp_canvas})$^{\ddagger}$ & --- (probe) \\
\bottomrule
\end{tabular}
\caption{\textbf{Which prompt each experiment used.} $^{\dagger}$Translation reuses the canvas system prompt unchanged but passes the prose description as the user message directly, with no \texttt{Generate a 24x24 pixel art mosaic:} wrapper, since the prose \emph{is} the full specification. $^{\ddagger}$The probes condition on exactly the same chat-templated input used for real generation, so the representation being read is the one generation actually starts from. Judge~1 (Qwen2.5-VL-7B) and Judge~2 (InternVL3-8B) use byte-identical prompts; they differ only in the model.}
\label{tab:prompt_matrix}
\end{table}

\subsubsection{Canvas generation system prompt}
\label{app:sp_canvas}
Used for the Layout task, the Translation gate, and as the conditioning for every probe. The user message is \texttt{Generate a 24x24 pixel art mosaic: \{prompt\}}.

\begin{lstlisting}[style=prompt]
You are a pixel art generator. You write Python code using a 24x24 canvas API to create pixel art images from text descriptions.

CANVAS API (already available — do NOT import anything):
  canvas.fill(color)                         # fill entire canvas with color
  canvas.set_pixel(r, c, color)              # set single pixel at row r, col c
  canvas.rect(r, c, h, w, color)             # filled rectangle: top-left (r,c), h rows tall, w cols wide
  canvas.circle(cr, cc, radius, color)       # filled circle centred at (cr,cc)
  canvas.line(r1, c1, r2, c2, color)         # 1-pixel line between two points
  canvas.poly([(r,c), ...], color)           # filled polygon

COLORS:
  Use standard hex color values. The available color space is the full RGB color wheel:
    '#RRGGBB'      24-bit RGB, preferred
    '#RRGGBBAA'    32-bit RGBA, alpha is accepted but ignored by the RGB rasterizer
    0xRRGGBB       packed RGB integer
    0xAARRGGBB     packed 32-bit integer, alpha is accepted but ignored
    (r, g, b)      RGB tuple with channels 0-255
  Prefer hex strings such as '#ffffff', '#141414', '#d23232', '#87ceeb', '#d2d2d2'.

COORDINATE SYSTEM:
  Canvas is 24 rows × 24 cols. Row 0 = top, row 23 = bottom. Col 0 = left, col 23 = right.
  Centre of canvas is approximately (11,11) to (12,12).

NAMED COLORS (allowed for convenience, but hex strings are preferred):
  white, black, red, blue, green, yellow, orange, purple, teal, pink,
  navy, brown, gray, grey, light_gray, dark_gray, dark_grey, gold,
  light_blue, dark_blue, dark_navy, sky_blue, dark_green, dark_red,
  coral, beige, light_brown, dark_brown, dark_purple, dark_orange

RULES:
  1. Output ONLY valid Python code. No markdown fences, no comments, no explanations.
  2. No import statements. No print statements. No variable names starting with __.
  3. Always set the background explicitly as the first call canvas.fill('<appropriate_color>') — before drawing anything else. Choose the background color based on the prompt.
  4. Stay within bounds: rows 0–23, cols 0–23.
  5. Keep code concise — use loops where appropriate.
\end{lstlisting}

\subsubsection{SVG generation system prompt (Exp.~2)}
\label{app:sp_svg}
The user message is \texttt{Generate an SVG image: \{prompt\}}.

\begin{lstlisting}[style=prompt]
You are an SVG image generator. You write raw SVG markup to create images from text descriptions.

OUTPUT FORMAT:
  A single <svg> document, viewBox="0 0 512 512". Use standard SVG shape elements only:
    <rect x="." y="." width="." height="." fill="."/>
    <circle cx="." cy="." r="." fill="."/>
    <ellipse cx="." cy="." rx="." ry="." fill="."/>
    <line x1="." y1="." x2="." y2="." stroke="." stroke-width="."/>
    <polygon points="x1,y1 x2,y2 ..." fill="."/>
    <polyline points="x1,y1 x2,y2 ..." fill="none" stroke="."/>
    <path d="..." fill="."/>
    <g transform="..."> ... </g>  (grouping and simple transforms only)

REPEATING / TILED PATTERNS:
  For anything that repeats many times across the canvas (a brick wall, a checkerboard, a fence,
  a row of many identical items), do NOT list every repeat individually -- define ONE tile inside
  <defs><pattern>...</pattern></defs> and fill a single rect with it, the SVG equivalent of a loop:
    <defs>
      <pattern id="tile" x="0" y="0" width="W" height="H" patternUnits="userSpaceOnUse">
        <!-- shapes for ONE tile, positioned within a local 0,0 to W,H box -->
      </pattern>
    </defs>
    <rect x="0" y="0" width="512" height="512" fill="url(#tile)"/>
  You can also reuse a single shape you defined once in <defs> with <use href="#id" x="." y="."/>
  instead of redrawing its full geometry every time.

COLORS:
  Use standard hex color values ('#RRGGBB') or standard CSS color names
  (white, black, red, blue, green, yellow, orange, purple, teal, pink, navy,
  brown, gray, gold, skyblue, darkblue, darkgreen, darkred, coral, beige).

COORDINATE SYSTEM:
  The canvas is a 512x512 square, viewBox="0 0 512 512". (0,0) is top-left,
  (512,512) is bottom-right. Centre of the canvas is (256,256).

RULES:
  1. Output ONLY the SVG markup, starting with <svg and ending with </svg>. No markdown
     fences, no comments, no explanations, no text before or after the SVG.
  2. No <script> elements, no external references (no <image> tags, no href/xlink:href
     pointing at a URL), no <style> with @import. Everything must be self-contained.
     Local references within the same document, like href="#tile" for <use> or <pattern>,
     are fine and expected -- only URLs are disallowed.
  3. Always give the image an explicit background: the first child should be a
     <rect x="0" y="0" width="512" height="512" fill="..."/> covering the whole canvas,
     unless the description calls for a transparent/white background (SVG's default).
  4. Stay within the 512x512 canvas.
  5. Keep the markup concise.
\end{lstlisting}

\subsubsection{Judge system prompt (Exp.~1, both judges)}
\label{app:sp_judge}
Judge~1 and Judge~2 receive byte-identical system prompts and rubrics; they differ only in the underlying VLM.

\begin{lstlisting}[style=prompt]
You are evaluating pixel art images generated by an AI.
Each image is a 24×24 grid of colored pixels shown at higher resolution.
Score how well the art matches the description. Use integers 0–5 only:
  0=no match, 1=minimal, 2=partial, 3=reasonable, 4=good, 5=excellent
Respond with ONLY valid JSON, no other text.
\end{lstlisting}

\subsubsection{Medium-neutral judge system prompt (Exp.~2)}
\label{app:sp_generic}
Both arms of the medium ablation --- the SVG generations and a re-scored canvas baseline --- are judged through this prompt, so any score gap is attributable to the medium and not to which system prompt the judge saw. It differs from \ref{app:sp_judge} in exactly two respects: ``pixel art images'' becomes ``images'', and the sentence asserting a $24{\times}24$ pixel grid --- true of the canvas medium, false of natively-rasterised SVG --- is removed.

\begin{lstlisting}[style=prompt]
You are evaluating images generated by an AI from a text description.
Score how well the image matches the description. Use integers 0-5 only:
  0=no match, 1=minimal, 2=partial, 3=reasonable, 4=good, 5=excellent
Respond with ONLY valid JSON, no other text.
\end{lstlisting}

\subsubsection{Scoring rubric (user message, all judges)}
\label{app:sp_rubric}
Imported unchanged by every judging path, so the five dimensions are worded identically across judges and across media. Note that this rubric retains the phrase ``pixel-art image''; the medium-neutral change in \ref{app:sp_generic} applies to the system prompt only. Both arms of Exp.~2 receive this identical rubric, so the contrast remains symmetric, and if the residual framing has any effect it disadvantages the SVG arm and is therefore conservative with respect to the reported gap.

\begin{lstlisting}[style=prompt]
Description: "{prompt_text}"

Score how well this pixel-art image matches the description.
Use integers 0–5 only: 0=no match, 1=minimal, 2=partial, 3=reasonable, 4=good, 5=excellent

Score these five dimensions:
- prompt_fidelity: overall match to the description
- shape_accuracy: shapes rendered correctly
- color_accuracy: colors match what was requested
- spatial_accuracy: spatial relationships correct (position, inside, above, etc.)
- completeness: all required elements present

Respond as JSON only: {"prompt_fidelity": X, "shape_accuracy": X, "color_accuracy": X, "spatial_accuracy": X, "completeness": X, "reasoning": "one sentence"}
\end{lstlisting}

\subsection{Example synthetic prompts and mosaic outputs}
\label{synthetic_mosaic_examples}
\begin{figure*}[hbt!]
\begin{center}
\includegraphics[width=0.96\linewidth]{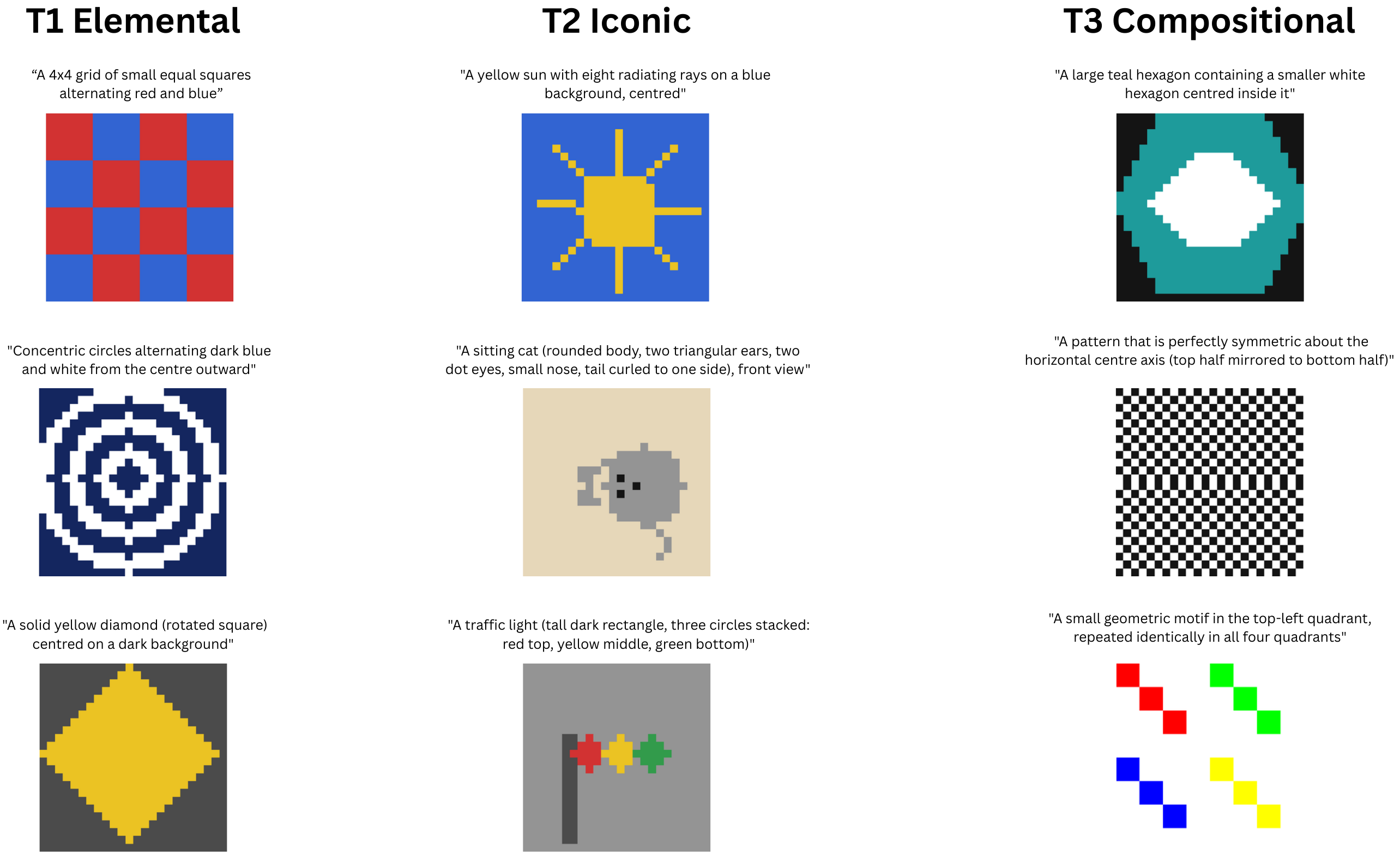}
\end{center}
\caption{\textbf{Example mosaics from synthetic prompts}}
\label{fig:synthetic_mosaics}
\end{figure*}

See Fig.~\ref{fig:synthetic_mosaics}

\subsection{Example human prompts and mosaic outputs}
\label{human_mosaic_examples}
\begin{figure*}[hbt!]
\begin{center}
\includegraphics[width=0.96\linewidth]{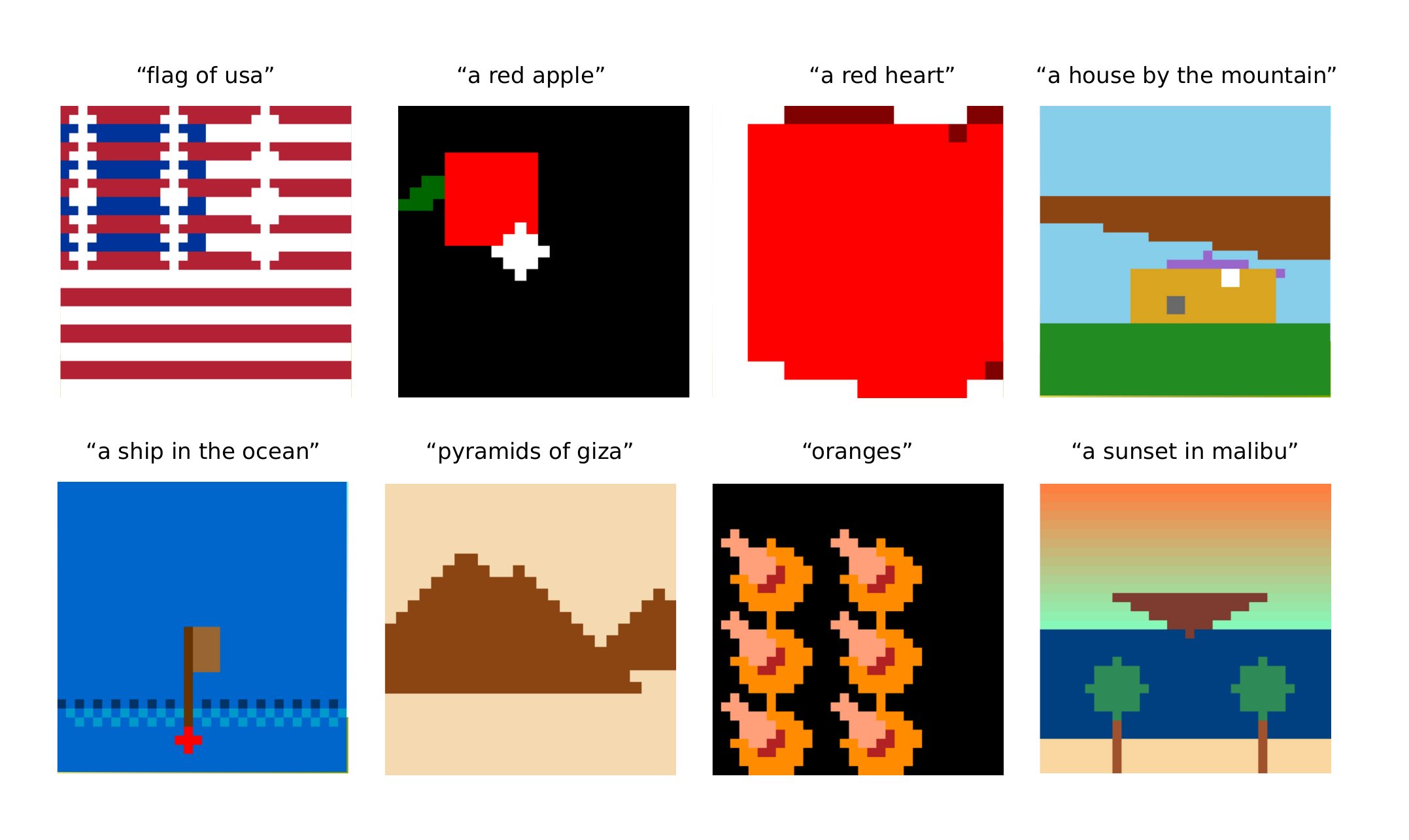}
\end{center}
\caption{\textbf{Example mosaics from human prompts}}
\label{fig:human_mosaics}
\end{figure*}

See Fig.~\ref{fig:human_mosaics}

\section{The Translation Task (Control)}
\label{app:translation_section}

\subsection{Symbolic metric details}
\label{app:metric}
Part assignment uses overlap ratios $|g_j\cap R_i|/|g_j|$ with the exact-area machinery of Eq.~\ref{eq:piou}; ties go to the \emph{smaller} part (Sec.~\ref{sec:translation_design}); \texttt{set\_pixel} is treated as a $1$-cell square. Background-assigned area enters a separate over-painting penalty reported beside PIoU.

\subsection{Translation stress-test details}
\label{app:translation_details}

Full detail behind the translation task: why it is not just a lenient metric, how the 100-reference difficulty ladder and 13-icon set were built and what they show, and what predicts the rare cases where translation does fail. \textbf{Everything in this appendix is scoped to the original two-model pilot} (Qwen2.5-Coder-14B and a 7B pilot model, the latter not part of the eight-model roster reported in Sec.~\ref{sec:experiments}); none of it was re-run for the other six models.

\begin{table}[hbt!]
\centering
\small
\setlength{\tabcolsep}{4pt}
\renewcommand{\arraystretch}{0.95}
\begin{tabular}{@{}lcccc@{}}
\toprule
Model & Mean & PIoU$_{\text{pen}}$ & \% $\geq$ 0.6 & \% $=$ 1.0 \\
\midrule
Qwen2.5-Coder 32B & 0.994 & 0.998 & 99.5 & 97.3 \\
Qwen2.5-Coder 14B & 0.995 & 0.995 & 99.3 & 97.9 \\
Gemma~2 27B & 0.992 & 0.999 & 99.3 & 98.8 \\
Gemma~2 9B & 0.980 & 0.986 & 97.4 & 96.8 \\
GLM-4 32B & 0.993 & 0.999 & 99.3 & 98.9 \\
GLM-4 9B & 0.991 & 0.996 & 99.1 & 98.8 \\
CodeLlama~34B & 0.984 & 0.994 & 98.3 & 97.6 \\
Llama~3.1 8B & 0.929 & 0.945 & 92.9 & 86.8 \\
\bottomrule
\end{tabular}
\caption{\textbf{Translation task results, all eight models.} Mean PIoU over the 145-reference set. The mean sits below the median (median=1.0) only because every model has at least one reference it fails outright. Unlike the rest of this appendix, this table covers the full eight-model roster.}
\label{tab:translation_headline}
\end{table}

\begin{figure}[h]
\begin{center}
\includegraphics[width=0.95\linewidth]{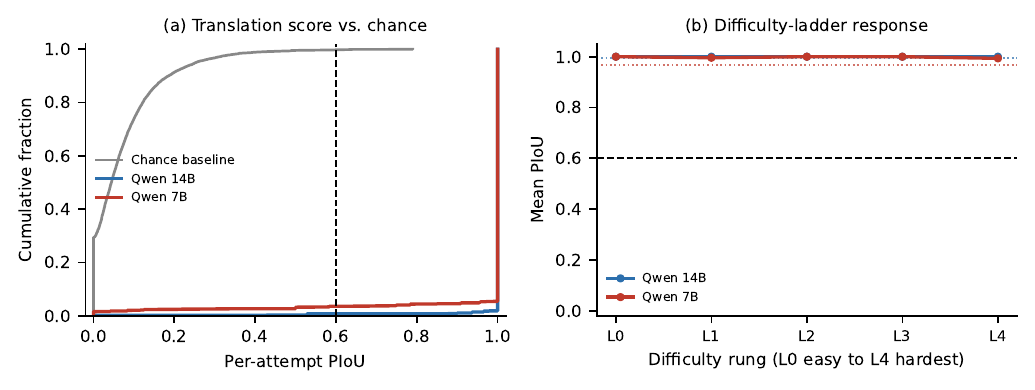}
\end{center}
%\vspace{-3mm}
   \caption{\textbf{Translation gate.} (a)~Real per-attempt scores against a chance-level null baseline. (b)~The 100-reference stress-test ladder: mean PIoU by rung, both models, against each model's first-round mean (dotted).}
\label{fig:translation_gate}
\end{figure}

\textbf{Not just a lenient metric.} A model could exceed the threshold for passing the task simply because the metric is easy to satisfy by chance. We rule this out with a null baseline: 100 type-and-count-matched random placements per reference (145 references $\times$ 100 seeds, 14,500 placements, same scoring pipeline). Random placement scores a mean PIoU of 0.072, with 95\% of random attempts below 0.253 --- far below both the threshold and the real scores (Fig.~\ref{fig:translation_gate}a).

\textbf{Over-paint and color cost.} Penalising every attempt for unrequested extra shapes barely moves either score: 0.9951 to 0.9935 for 14B, 0.9679 to 0.9636 for 7B; only 0.4\% of 14B's attempts and 2.6\% of 7B's had any meaningful over-paint. Requiring the assigned color to match as well as the shape costs 0.033 for 14B (0.995 to 0.962) and 0.036 for 7B (0.968 to 0.932) --- real, but not enough to change the outcome.

\textbf{The difficulty ladder.} Because the 145 references were selected from pictures the models had already drawn successfully, we built 100 new ones procedurally, so that a correct transcription exists by construction rather than by prior success. They span four difficulty axes independently, in five rungs from easy (matched to the first round) to hardest (every axis pushed at once): part count (3 to 16), minimum feature size relative to the canvas (0.15 to 0.045, the resolution floor), the gap left between adjacent shapes (0.15 to 0), and aspect ratio of elongated shapes (2:1 to 10:1); 20 references per rung. Before generating anything we pre-registered two outcomes: scores staying within 0.05 of the first round's mean at every rung, or dropping by at least 0.15 at the hardest rung.

Fig.~\ref{fig:translation_gate}b shows the result: scores stay flat to slightly \emph{higher} at every rung, for both models. Per-rung means: 14B scores 1.000 at every rung, L0 through L4; 7B scores 1.000, 0.996, 1.000, 1.000, 0.993 (mean 0.998 across the five rungs), against first-round means of 0.995 and 0.968.

This is the honest direction to state, not the flattering one. The ladder's four axes are all axes of \emph{geometric} difficulty, and translation is a transcription task: a model copying a stated radius never has to judge whether that radius is small. The axis most likely to matter for transcription --- periodicity, the repeating grids and tilings that invite a generative loop instead of item-by-item placement --- was not built into the ladder and remains untested. It is also the most plausible reading of an unrelated pattern already in this data: the 7B model's weakest Layout tier is T1 Elemental (Sec.~\ref{sec:exp1}), not the harder tiers, with its worst subcategories being Patterns (1C) and Symmetry (3E).

\begin{figure}[h]
\begin{center}
\includegraphics[width=0.85\linewidth]{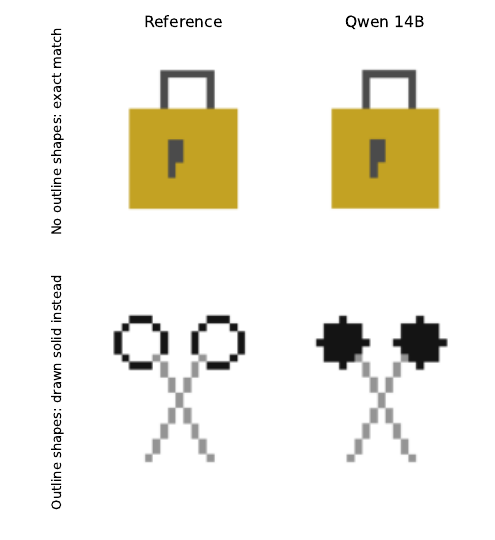}
\end{center}
%\vspace{-3mm}
   \caption{\textbf{Reference versus generation for two icons.} Top: no outline shapes, an exact match. Bottom: two outline-ring handles are drawn as solid discs instead.}
\label{fig:translation_icon}
\end{figure}

\textbf{Icon set.} 13 hand-drawn references, curated by a human rater for recognisability: padlock, key, teapot, bicycle, umbrella, wine glass, hourglass, house, envelope, scissors, hammer, coffee mug, camera. Across all 13, mean score fell to 0.90 --- the one real drop anywhere in translation, and it traces entirely to a documented gap in what the models were taught rather than to reasoning. Five icons use an unfilled outline shape (scissors, camera, bicycle, key, coffee mug); the remaining eight do not. Splitting them this way explains the entire gap: the eight with no outline shape score a perfect 1.000 for both models, while the five that do score exactly 0.731 for both models, to three decimal places --- precisely what every outline shape scoring zero and everything else scoring perfectly would produce. The drawing tool supports an unfilled shape, but the written instructions given to both models describe only the filled form (Fig.~\ref{fig:translation_icon}). See also Sec.~\ref{sec:exp1}'s note on Layout subcategory 1B, which needs the same visual effect and finds a working substitute using only documented primitives.

\textbf{What predicts a rare translation miss.} Misses are rare, but when they happen they are predicted far better by the model's own token-level uncertainty while writing the code than by target-picture complexity. Recast as logistic regression predicting whether an attempt falls below $\tau_{\text{trans}}$ from mean token-level entropy on that attempt (rather than a Spearman correlation on a variable degenerate at 1.0, which is dominated by tie handling): the odds ratio is 4.8 (95\% CI 3.1--7.4) for 14B and 2.6 (95\% CI 1.9--3.6) for 7B per one-unit increase in mean entropy, while picture complexity (part count) gives an odds ratio indistinguishable from 1 for both models. Translation misses are principled, not random noise.

\textbf{Gate recomputation.} Median PIoU stays at 1.000 for both models whether computed on the first-round references alone, the stress-test ladder alone, or all 245 pooled.

\section{The Layout Task \& Evaluation (Target)}
\label{app:layout_section}

\subsection{VLM judge details}
\label{app:judge}

\textbf{Rubric.} Five dimensions scored 0--5 (integers only): (1)~\emph{prompt fidelity}, (2)~\emph{shape accuracy}, (3)~\emph{color accuracy}, (4)~\emph{spatial accuracy}, (5)~\emph{completeness}. Composite \texttt{vlm\_score} is the mean of the five dimensions; pass $\geq 3.0$.

\textbf{Per-dimension scores under Judge~1 (all tiers pooled, all eight models):}

\begin{center}
\small
\setlength{\tabcolsep}{3pt}
\begin{tabular}{@{}lccccc@{}}
\toprule
Model & Fid. & Shape & Color & Spatial & Compl. \\
\midrule
GLM-4 32B         & 3.56 & 3.57 & 3.80 & 3.60 & 3.57 \\
Qwen2.5-Coder 32B & 3.37 & 3.37 & 3.69 & 3.40 & 3.37 \\
Qwen2.5-Coder 14B & 3.24 & 3.23 & 3.51 & 3.29 & 3.24 \\
Gemma~2 27B       & 2.83 & 2.83 & 3.27 & 2.87 & 2.84 \\
GLM-4 9B          & 2.81 & 2.82 & 3.21 & 2.85 & 2.81 \\
Gemma~2 9B        & 2.32 & 2.32 & 2.79 & 2.35 & 2.32 \\
Llama~3.1 8B      & 1.94 & 1.95 & 2.38 & 1.96 & 1.95 \\
CodeLlama~34B     & 1.65 & 1.64 & 2.19 & 1.70 & 1.65 \\
\midrule
Pooled            & 2.72 & 2.72 & 3.11 & 2.76 & 2.72 \\
\bottomrule
\end{tabular}
\end{center}

Color accuracy is the clear standout highest dimension for every model, pooled and individually. The other four dimensions track each other closely per model (spatial accuracy marginally highest of the four, pooled), with no dimension consistently weakest across the roster.

\begin{figure*}[hbt]
\begin{center}
\includegraphics[width=0.98\linewidth]{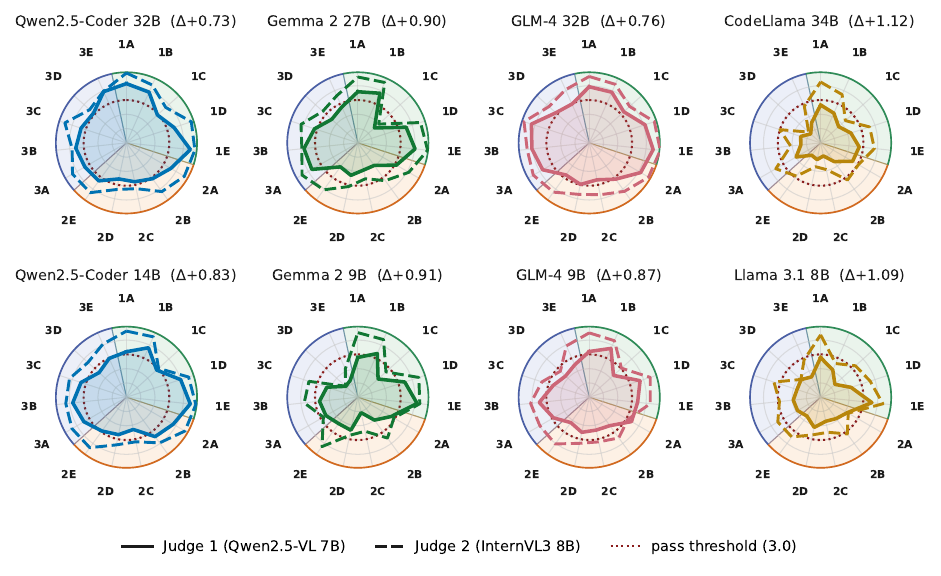}
\end{center}
%\vspace{-3mm}
   \caption{\textbf{The two VLM judges compared per model and subcategory (15 subcategories $\times$ 8 models).} One panel per model, with both judges on the same axes: solid is Judge~1 (Qwen2.5-VL~7B, filled) and dashed is Judge~2 (InternVL3~8B). \emph{Line style here denotes judge, not model size} --- the model is fixed within a panel. Each panel title gives that model's mean J2$-$J1 gap. Radial axis is the score, 0 at the centre to 5 at the rim; the dotted circle is the pass threshold (3.0). Background color separates the tiers: T1 -- Green; T2 -- Orange; T3 -- Blue. The two profiles are near-concentric for every model: J2 sits uniformly outside J1, by $+$0.73 (Qwen2.5-Coder~32B) to $+$1.12 (CodeLlama~34B), while the shape of each profile --- which subcategories a model is relatively strong or weak on --- is preserved. This is the per-subcategory form of the inter-judge relationship reported in Sec.~\ref{sec:exp1}: the judges disagree on absolute level but agree on ordering, with a model-level Spearman $\rho = 1.000$.}
\label{fig:subcat_radar_judges}
\end{figure*}

\textbf{Judge 1.} Qwen2.5-VL-7B-Instruct; BF16; $\sim$1.7~s/image; 12,932 valid images scored across all eight models (Sec.~\ref{sec:experiments}).

\textbf{Judge 2.} OpenGVLab/InternVL3-8B \cite{chen2025internvl3}; BF16; \texttt{trust\_remote\_code=True}; $448{\times}448$ BICUBIC resize; $\sim$2~s/image; 12,926 scored, six fewer than Judge~1 (six images, spread across five models, Judge~2 could not process; five cluster on a single prompt, a clock face.

\begin{figure}[hbt]
\begin{center}
\includegraphics[width=0.95\linewidth]{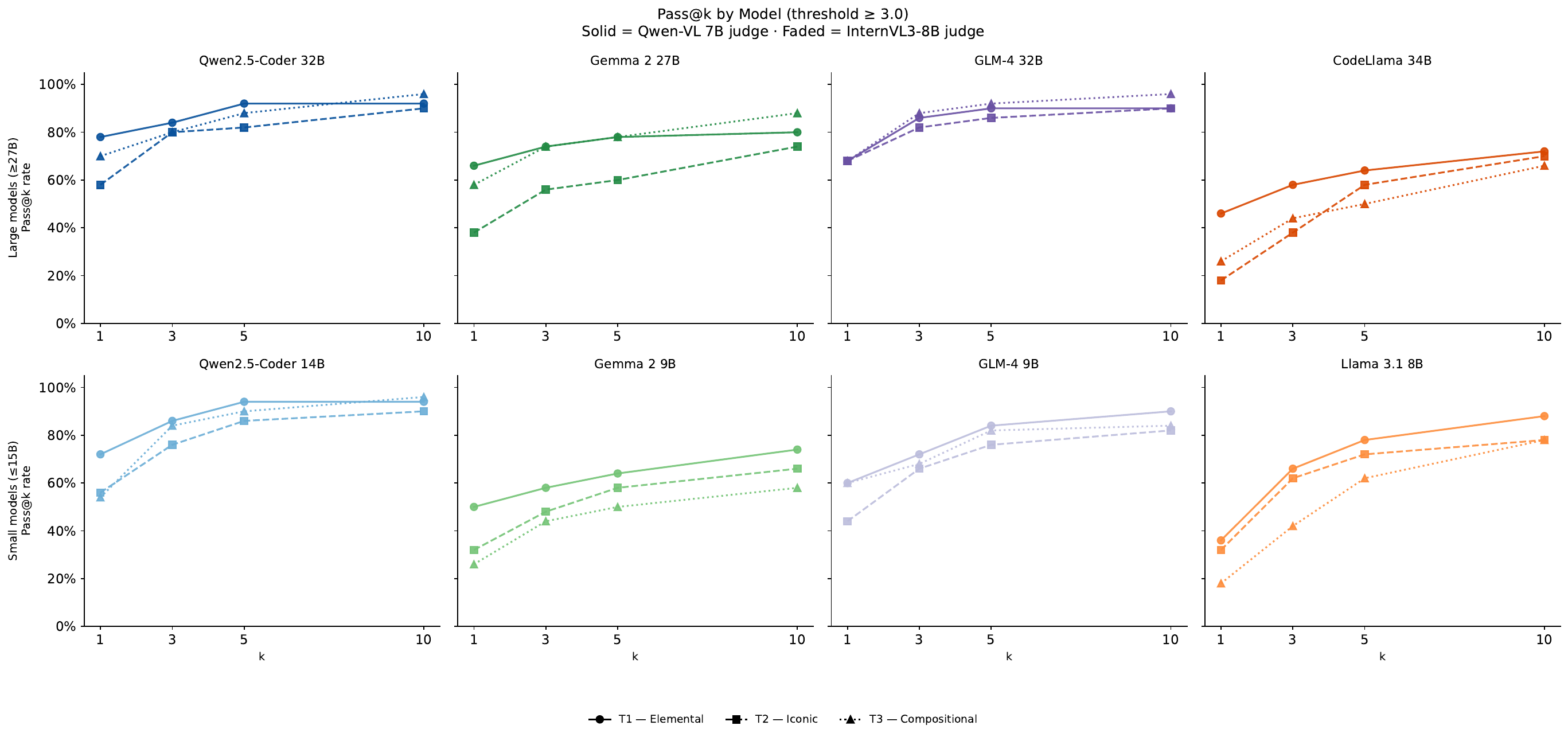}
\end{center}
%\vspace{-3mm}
   \caption{\textbf{Pass@$k$ ($k \in \{1,3,5,10\}$; pass $=$ valid $\wedge$ non-trivial $\wedge$ vlm\_score $\geq 3.0$), all eight models.} Under J1, pass@10 ranges from 0.66 (Gemma~2~9B) to 0.93 (Qwen2.5-Coder~14B) overall, and reaches 0.96 at T3 for the three strongest models.}
\label{fig:passatk}
\end{figure}

\subsection{Human-validation materials and results}
\label{app:human}
Referenced from Sec.~\ref{sec:limitations}. A script builds a single fixed pool of 350 pairs from the existing VLM-scored generations: 240 pairs stratified by $|\Delta J1|$ into four bins (0--0.5, 0.5--1, 1--2, 2+), 60 cross-model pairs (same prompt, 14B vs.\ 7B), and 35 cross-tier pairs (T2 vs.\ T3), plus 15 blank-canvas catch trials for quality control, allocated across prompts round-robin so a handful of prompts cannot crowd out the rest (149 of 150 Layout prompts appear at least once). Every annotator sees this identical 350-pair set; only presentation order and left/right side are randomised per annotator, so responses stay directly comparable pair-for-pair. A self-contained HTML page presents one pair at a time (arrow-key or click controls, explicit ``both valid'' and ``neither valid'' options, and an undo step), with no backend and no external dependency, and exports raw responses as JSON.

\textbf{Two annotators with full overlap.} Two annotators independently judged all 350 pairs. With two raters a pair has a verdict only where both choose the same option; disagreements are dropped rather than broken arbitrarily, so all rates below are conditioned on agreement. The annotators agree on 247 of 350 pairs (70.6\%), Krippendorff's $\alpha = 0.60$. Of the 247 agreed pairs, 126 carry a directional verdict and 121 are agreed ties (``both valid'' or ``neither valid''). Restricting to agreed pairs where the judge also separates the two images, the human verdict matches Judge~1 on 82.4\% (61/74) and Judge~2 on 76.7\% (46/60), well above chance and above the corresponding single-annotator rates (74.8\% and 71.3\%), as expected once ambiguous pairs are filtered by requiring agreement.

\textbf{Ties track the judges' own score gap.} The agreed-tie rate falls monotonically with $|\Delta J1|$: 66.7\% at $[0,0.5]$, 50.0\% at $[0.5,1]$ and $[1,2]$, and 34.1\% at $[2,5]$. Two annotators independently declining to separate a pair precisely where the judges score it as close is evidence that the gap reflects visual similarity rather than judge noise.

\textbf{Secondary comparisons remain underpowered.} On agreed pairs, the cross-model 14B-over-7B win rate is 76.2\% (95\% CI $[52.8, 91.8]$, $n{=}21$) and the cross-tier T2 win rate is 28.6\% (CI $[3.7, 71.0]$, $n{=}7$). Requiring agreement removes most of these pairs, so both intervals are wide and the cross-tier interval spans the no-preference point; neither is a confirmed replication of the corresponding automated result.

% Unnessary 
% \textbf{Catch trials and a disclosed deviation.} Catch-trial accuracy was 14/15 (93.3\%) for the first annotator and 11/15 (73.3\%) for the second, the latter below the 90\% exclusion threshold specified in advance. All four of that annotator's errors were the same response, ``neither valid'', on catch trials whose non-blank image is itself a poor match for its prompt, at response times indistinguishable from the rest of the session; neither annotator ever selected the blank canvas or marked it ``both valid'' (15/15 each), which is the behaviour the catch trial is designed to detect. We report the two-annotator result with this deviation stated rather than discarding a complete annotation set on a check that conflates inattention with a judgement about image quality. Under the threshold as originally specified, the primary result reverts to the single-annotator rates given above.

\textbf{Scope.} $\alpha = 0.60$ indicates moderate agreement, below the 0.667 conventionally treated as a floor for tentative conclusions, so these rates support the claim that both judges track human preference well above chance, not that human raters agree strongly with one another. A larger multi-annotator run in future work would improve the study but is out of scope for this work.

\section{Deep-Dive Experimental Details}
\label{app:deep_dive}

\subsection{Exp.~2: Medium Ablation Details}

\begin{table}[hbt!]
\centering\footnotesize
\setlength{\tabcolsep}{4pt}
\begin{tabular}{@{}lccc@{}}
\toprule
Model & Native $\Delta$ & Resolution-matched $\Delta$ & 95\% CI \\
\midrule
CodeLlama 34B      & $+$0.84 & $+$0.76 & $[0.56, 0.97]$ \\
Gemma 2 9B         & $+$0.64 & $+$0.56 & $[0.34, 0.83]$ \\
Llama 3.1 8B       & $+$0.54 & $+$0.51 & $[0.32, 0.68]$ \\
GLM-4 9B           & $+$0.58 & $+$0.39 & $[0.22, 0.60]$ \\
Gemma 2 27B        & $+$0.38 & $+$0.26 & $[0.04, 0.50]$ \\
GLM-4 32B          & $+$0.33 & $+$0.15 & $[0.02, 0.35]$ \\
Qwen2.5-Coder 14B  & $+$0.23 & $+$0.15 & $[-0.05, 0.33]$ \\
Qwen2.5-Coder 32B  & $+$0.32 & $+$0.09 & $[-0.10, 0.27]$ \\
\midrule
Pooled             & $+$0.49 & $+$0.37 & $[0.26, 0.47]$ \\
\bottomrule
\end{tabular}
\caption{\textbf{Medium ablation, full values.} \emph{Native} is the raw canvas-vs-SVG comparison, confounded by SVG's anti-aliased native rendering. \emph{Resolution-matched} re-renders every SVG generation at $24{\times}24$ from its vector source before rescoring, isolating the medium from rendering fidelity; this is the result reported in Sec.~\ref{sec:exp_medium}. Sorted by resolution-matched $\Delta$. CI is on the resolution-matched estimate, prompt-clustered bootstrap.}
\label{tab:svg_corrected}
\end{table}

\subsection{Exp.~3: full detail}
\label{app:exp3_appendix}

Referenced from Sec.~\ref{sec:exp3}. Exp.~3 has two stages: a representational stage on what pre-generation activations encode, and a causal stage on whether the model uses that information while drawing.

\subsubsection{Stage 1: what the pre-generation residual stream encodes}

\textbf{Why a single reference generation is not an identifiable target.} The first version of this experiment decoded the occupancy of one reference generation per prompt. That target is not well posed: an underspecified Layout prompt admits many valid layouts, so one generation samples the distribution rather than defining it. All 145 references also originate from two models (83 from Qwen2.5-Coder~14B, 62 from a 7B pilot), so seven of the eight roster models are probed against another model's output. We bound this directly with an oracle no probe can beat: a model's own perfect consensus occupancy predicts the reference at 0.688 on average (per-model range 0.632--0.774), against a text baseline of 0.625. Four of eight models retain under 0.05 of headroom (Llama~3.1~8B 0.030, CodeLlama~34B 0.039, GLM-4~9B 0.043) and Gemma~2~9B retains 0.007. A null measured inside that band constrains the target, not the representation.

\textbf{Activations.} We run a single forward pass (\texttt{output\_hidden\_states=True}, no generation) over each of the 145 real Layout prompts, using the same system prompt, user prompt, and chat template as real generation elsewhere in this paper. We capture hidden states at every layer (33--65 including the embedding layer, from Llama~3.1~8B's 32 transformer layers to Qwen2.5-Coder~32B's 64) and at three token positions rather than one: the last template token, the last content token, and the mean over content tokens. The original single-position design used template boilerplate, identical across prompts under some chat templates. Extraction is bitwise-deterministic across repeated runs for every model.

\textbf{Target.} The model's own consensus occupancy: for each prompt, the per-cell mean of the $6{\times}6$ occupancy grids over that model's valid generations. The target is a continuous occupancy probability, not one binary realisation, and it belongs to the model being probed. Prompts with too few valid generations are dropped, so $n$ ranges 138--145 across models.

\textbf{Probes and nested cross-validation.} Standardisation, then PCA, then ridge regression. Layer, token position, PCA dimensionality, and ridge penalty are selected in an inner loop on training folds only; the outer loop reports out-of-fold pooled $R^2$. The original design selected the best layer by \texttt{idxmax} over the reported test folds, which is selection on test data. The TF-IDF text baseline is one pre-registered configuration, refit inside each training fold rather than taken as a maximum over eight variants fitted on all 145 prompts, so neither arm is advantaged by selection \cite{hewitt2019control}. Seeding and reproducibility follow the discipline in Appendix~\ref{app:stats}.

\textbf{Result and controls.} Decodability exceeds the text baseline for all eight models, every interval excluding zero and every model significant after Benjamini-Hochberg correction: mean $\Delta R^2 = {+}0.164$, from ${+}0.130$ (Qwen2.5-Coder~14B) to ${+}0.205$ (Gemma~2~9B), all $p_{\mathrm{FDR}} \leq 0.0004$, prompt-clustered bootstrap with 10{,}000 resamples. Two controls run alongside. Permuting the prompt-to-activation mapping gives $R^2$ between $-0.036$ and ${+}0.002$, effectively zero for every model. Probing the embedding layer gives a mean $R^2$ of 0.132 against the text baseline's 0.149: expected, since layer-0 activations approximate a bag-of-tokens representation, so a control that matches the text baseline behaves correctly.

\textbf{Cross-model selectivity.} Each model's probe is refit against every other model's consensus target, giving an $8{\times}8$ matrix. A representation encoding \emph{this} model's layout should decode its own target best. It does not. Own-source is the best of the eight for one model only (Gemma~2~9B); the mean own-minus-others gap is ${+}0.015$, and for GLM-4~9B it is negative ($-0.007$). Eight models decode one another's layouts about as well as their own -- an early sign that the probe reads a generic signal.

\textbf{Decomposition.} On the two strongest large models we split the target into a \emph{shared} component, the leave-one-model-out consensus of the other seven models, and a \emph{specific} component, the residual after subtracting it. Selected layer and position are layer 43 / mean-content for GLM-4~32B and layers 23--26 / mean-content for Gemma~2~27B. Table~\ref{tab:decomposition} reports the result. The shared component decodes at $R^2 = 0.550$ and $0.573$ against text baselines of $0.289$ and $0.293$; the specific component reaches $0.005$ and $-0.005$. We report the specific component as an absolute $R^2$, not as a gap over text: its text baselines are themselves slightly negative, so the \emph{difference} is nominally non-zero (${+}0.019$, raw $p = 0.058$; ${+}0.024$, raw $p = 0.020$) while the probe explains no variance in absolute terms. A gap over a baseline that is below zero is not evidence of decoding.

\textbf{Reliability ceiling.} Split-half over each model's generations with Spearman-Brown correction bounds what any probe could reach: 0.955 and 0.978 for the own-layout target, 0.883 and 0.941 for the specific component. The observed values reach 33.3\% and 31.3\% of the ceiling for the own target, and 0.6\% and $-0.6$\% for the specific one. The specific-component null is a property of the representation, not of a noisy target.

\subsubsection{Stage 2: Behavioural causal test of whether the layout is used}

\textbf{Design.} Decodability is presence, not use \cite{li2023othello}. We test use behaviourally on the same two models. We truncate the model's own generations after its first drawing statement, then append the model's \emph{own} next statement in one of three forms: verbatim (\texttt{null\_asis}), translated so the shape lands where that generation never draws (\texttt{moved\_unused}, median displacement 9\,px), or translated onto cells it was going to fill later (\texttt{moved\_used}, median 4\,px). A fourth arm appends nothing. The model then continues from the identical prefix under the production sampling schedule, eight continuations per (item, arm): 731 items over 165 model-prompt combinations, 2{,}924 (item, arm) pairs, 26{,}316 scored continuations.

\textbf{Why the intervention moves the model's own statement.} Relocation is a textual digit substitution: new coordinates are spliced in at the numeric literals' own source offsets, so primitive type, size, color, argument count, whitespace, and quote style survive byte-for-byte. Stripping every digit from a moved statement and its null leaves identical strings for 1{,}462 of 1{,}462 relocations, so only position differs. An earlier design appended a synthesized rectangle instead; we abandoned it. Undisturbed, neither model ever tiles such rectangles (control rate 0.000), but appending one drove 17--73\% of continuations into a repetition loop -- induction on the statement's form, not a response to the picture. We log a repetition-loop rate in every run so this artifact cannot recur unnoticed.

\textbf{Metrics.} All measures are taken against the raster the model was handed, so the arm's own statement is never counted as the model's response to it. Union occupancy is monotone and cannot represent overpainting, so an occupancy-based avoidance metric is identically zero by construction. The headline measure is \texttt{drew\_anything}: whether the continuation emits any canvas call. It is a code-level measure, so unlike the pixel measures it cannot be moved by what happens to sit underneath a region.

\textbf{Confound stratification.} GLM-4~32B writes code comments in 21.3\% of its generations, Gemma~2~27B in 0.0\%. A comment announces the next primitive in plain language inside the context -- a look-ahead confound present in one model and absent in the other. The headline analysis runs on the comment-free subset; the inclusive result is a robustness row and does not change any conclusion.

\textbf{Results.} The clearest signal is in the stratum where the model had already finished its picture. Handed its own statement unmodified, it stops; handed the displaced version, it keeps drawing, from 43.8\% to 84.6\% for GLM-4~32B and from 2.5\% to 41.2\% for Gemma~2~27B ($\Delta = {+}0.415$, CI $[0.268, 0.564]$ and $\Delta = {+}0.387$, CI $[0.219, 0.563]$, both $p_{\mathrm{FDR}} = 0.0007$). The effect is graded by displacement size in both models independently: the 4\,px condition gives ${+}0.084$ (CI $[0.016, 0.168]$) and ${+}0.155$ (CI $[0.045, 0.290]$) against the 9\,px condition's ${+}0.415$ and ${+}0.387$. This dose-response is difficult to explain as an artefact of the intervention. Conditioning on the model drawing at all, so that ``draws more'' cannot pass for ``draws \emph{there}'', targeted restoration of the vacated location replicates in Gemma~2~27B only (${+}0.077$, $p = 0.031$ and ${+}0.107$, $p = 0.001$), not in GLM-4~32B, which trends the other way in the completed stratum ($-0.097$, $p = 0.002$). Validity is unchanged across arms, so the effect is not derailment. The models emit a byte-identical next statement under greedy decoding in 7.4--8.7\% of cases.

\textbf{Limits.} The pixel-change measures carry a mechanical confound: moving a shape changes what sits under each region, which is why the headline rests on the code-level measure and the targeting analysis is reported as mixed. The 4\,px condition is a weak manipulation -- its destination block overlaps the origin block 64--69\% of the time, so it functions as a low-dose condition rather than an independent one. The intervention perturbs the rendered state, not the representation, so it constrains behavior, not mechanism \cite{zhang2024towards}. Two models, 731 items: this is not a roster-wide claim.

\section{DINO Representational Analysis}
\label{app:dino_analysis}

\subsection{DINO extraction details}
\label{app:dino}
\textbf{Model.} \texttt{timm/vit\_small\_patch8\_224.dino} (ViT-S/8; 21M parameters; 384-d CLS token) loaded with \texttt{dynamic\_img\_size=True}, \texttt{img\_size=192}.

\textbf{Global coherence metrics.} $d_{\text{intra}}$ is the mean pairwise cosine distance within a prompt's $n$ samples, over whole-image CLS embeddings; $d_{\text{inter}}$ is the mean cosine distance from the prompt's centroid to every other prompt's centroid in the same tier $\times$ model stratum. Silhouette \cite{rousseeuw1987silhouettes} uses \texttt{scikit-learn} with \texttt{metric="cosine"}, one cluster per prompt.

\textbf{Local structure metric (Appendix~\ref{app:dino_consistency}).} Restricts DINO's last-block patch keys to a per-prompt foreground mask before computing self-similarity, so the measure is sensitive to spatial arrangement rather than overall appearance. The mask is built per prompt, independent of any single model or generation, by pooling a pixel-level (non-DINO) background/foreground classification across every generation of that prompt and keeping pixels foreground in $\geq$50\% of pooled generations; the 50\% threshold was tuned empirically against a handful of spot-checked prompts (25\% over-included via a union effect across generations; 60--75\% under-included). Structure distance between two generations is the Frobenius norm of the difference between their masked key-cosine self-similarity matrices, normalized by mask size. A small fraction of prompts describing several small, spatially diffuse elements produce an empty pooled mask (no pixel reaches 50\% agreement).

\textbf{UMAP.} A $\approx$2,000-image stratified subsample (roughly balanced across tiers); \texttt{umap-learn} \cite{mcinnes2018umap} with \texttt{metric="cosine"}, \texttt{n\_neighbors=20}, \texttt{min\_dist=0.1}.

\textbf{Attention extraction (Appendix~\ref{app:dino_attention}).} The last transformer block's self-attention is captured by injecting an instance-method override with \texttt{fused\_attn=False}, computing QKV directly, and saving the softmax attention weights. CLS-to-patch attention is the mean over the heads of $A[0,:,0,1{:}577]$, normalized to $[0,1]$ and reshaped to $24{\times}24$.

\subsection{DINO concept-consistency analysis}
\label{app:dino_consistency}
\label{app:dino_structure_appendix}
Appendix~\ref{app:translation_details} shows every model can express geometry it is simply given. This analysis asks a complementary question at the representation level: when a model is left to compose its own geometry, does it produce visually consistent pictures across repeated attempts at the same prompt, the signature of drawing from one stable internal picture rather than improvising anew each time? We test this using DINO ViT-S/8 embeddings \cite{dinov1} extracted from all 11,976 valid, non-trivial layout generations, and we deliberately keep two different notions of ``consistent'' apart: a \emph{global} one (CLS-token cosine distance, sensitive to overall semantic content: object identity, color, composition) and a \emph{local} one (masked, patch-key self-similarity distance restricted to the drawn foreground, sensitive specifically to spatial arrangement). A global measure alone cannot tell whether two attempts look alike because they share the same layout or merely the same palette. Therefore, we build the local measure to answer that more specific question.

\noindent\textbf{Preprocessing.}
Generated images ($512{\times}512$ nearest-neighbour upscales) are brought back to $24{\times}24$, then upscaled to $192{\times}192$ ($=24{\times}8$), so each DINO patch ($8{\times}8$ pixels) covers exactly one cell of the pixel-art canvas: a $24{\times}24$ patch grid, 576 patches. Appendix~\ref{app:dino} has the full extraction details, including how the local measure's per-prompt foreground mask is built.

\begin{figure}[hbt]
\begin{center}
\includegraphics[width=0.95\linewidth]{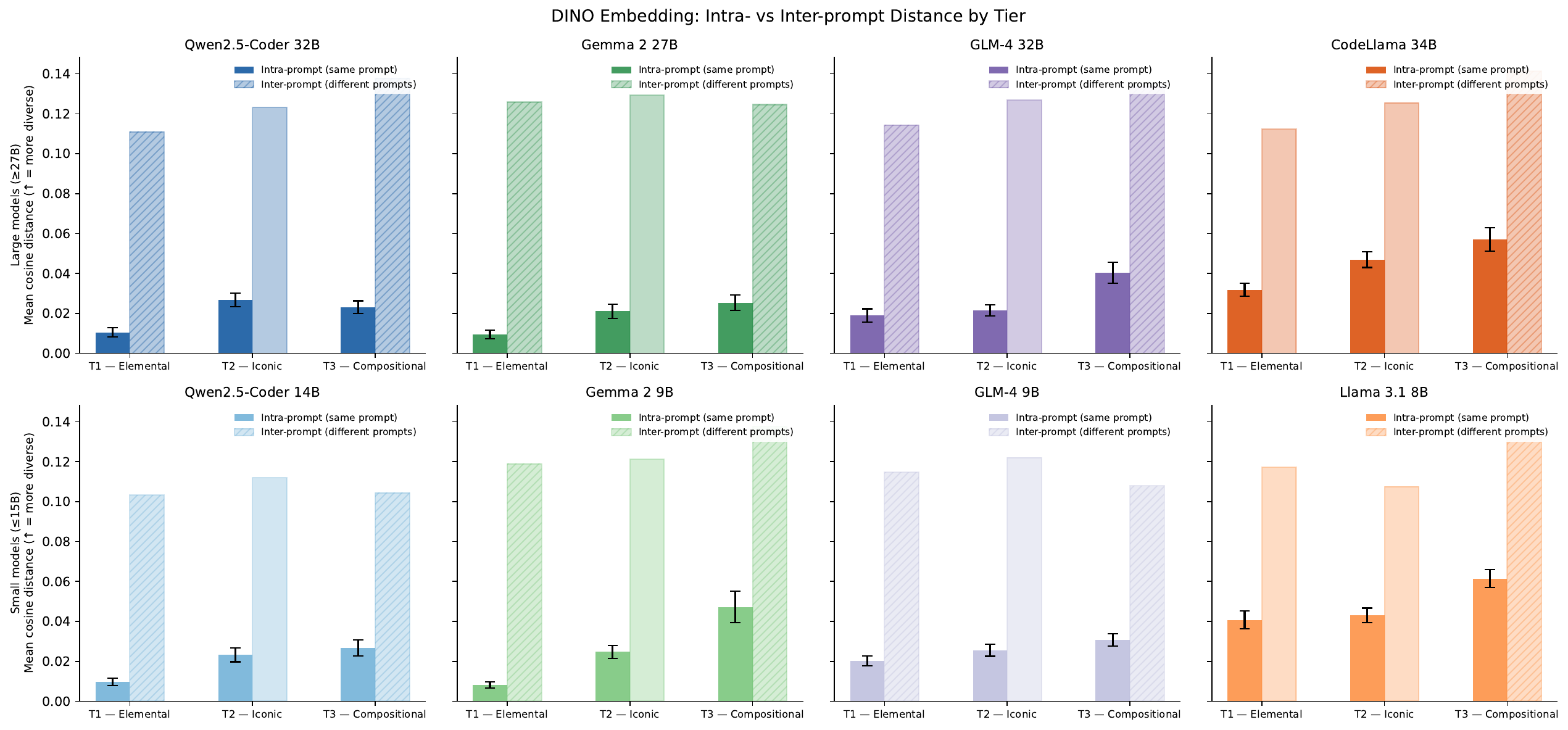}
\end{center}
%\vspace{-3mm}
   \caption{\textbf{Intra- vs.\ inter-prompt DINO cosine distances, all eight models.} GLM-4~32B and Qwen2.5-Coder~32B keep the largest gap at every tier; Llama~3.1~8B and CodeLlama~34B's intra-prompt distance approaches inter-prompt everywhere.}
\label{fig:dino_distances}
\end{figure}

\noindent\textbf{Global intra-prompt embedding coherence.}
For each prompt we compute the mean pairwise cosine distance within its 11 samples ($d_{\text{intra}}$) and the mean cosine distance from the prompt's centroid to every other prompt's centroid ($d_{\text{inter}}$), using CLS embeddings. Table~\ref{tab:dino_coherence} shows a clear model-by-tier interaction. GLM-4~32B and Qwen2.5-Coder~32B keep the tightest within-prompt clusters at every tier while Llama~3.1~8B and CodeLlama~34B are the most scattered, and are the only two models whose silhouette score \cite{rousseeuw1987silhouettes} goes negative at \emph{every} tier, a \emph{concept collapse} in which the model cannot keep distinct visual representations for different prompts apart, not confined to the harder tiers for these two. Fig.~\ref{fig:dino_distances} shows the underlying distance distributions.
% and Appendix~\ref{app:dino_structure_appendix} gives the full global-vs-local breakdown. % we are in this section already

\begin{table}[hbt]
\centering
\footnotesize
\setlength{\tabcolsep}{2.6pt}
\renewcommand{\arraystretch}{0.95}
\begin{tabular}{@{}lccc@{\hspace{5pt}}ccc@{}}
\toprule
& \multicolumn{3}{c}{$d_{\mathrm{intra}}/d_{\mathrm{inter}}$} & \multicolumn{3}{c}{Silhouette} \\
\cmidrule(r){2-4}\cmidrule(l){5-7}
Model & T1 & T2 & T3 & T1 & T2 & T3 \\
\midrule
GLM-4 32B          & 0.17 & 0.17 & 0.30 & \phantom{$-$}0.12 & \phantom{$-$}0.10 & $-$0.12 \\
Qwen2.5-Coder 32B  & 0.10 & 0.22 & 0.17 & \phantom{$-$}0.40 & $-$0.02 & \phantom{$-$}0.09 \\
Qwen2.5-Coder 14B  & 0.09 & 0.21 & 0.26 & \phantom{$-$}0.27 & \phantom{$-$}0.05 & \phantom{$-$}0.01 \\
Gemma~2 27B        & 0.08 & 0.16 & 0.20 & \phantom{$-$}0.38 & \phantom{$-$}0.20 & \phantom{$-$}0.09 \\
GLM-4 9B           & 0.18 & 0.21 & 0.28 & \phantom{$-$}0.09 & \phantom{$-$}0.06 & \phantom{$-$}0.06 \\
Gemma~2 9B         & 0.07 & 0.20 & 0.34 & \phantom{$-$}0.47 & \phantom{$-$}0.02 & $-$0.11 \\
Llama~3.1 8B       & 0.35 & 0.40 & 0.46 & $-$0.22 & $-$0.24 & $-$0.31 \\
CodeLlama~34B      & 0.28 & 0.37 & 0.40 & $-$0.12 & $-$0.39 & $-$0.30 \\
\bottomrule
\end{tabular}
\caption{\textbf{DINO intra-prompt coherence, all eight models.} Lower ratio means tighter within-prompt clusters. Llama~3.1~8B and CodeLlama~34B are the only models with negative silhouette at every tier.}
\label{tab:dino_coherence}
\end{table}

\subsection{Repeatability with DINO and the Entropy Confound}

\begin{figure}[hbt]
\begin{center}
\includegraphics[width=0.95\linewidth]{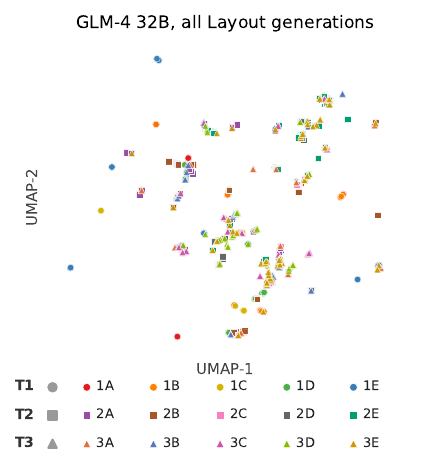}
\end{center}
%\vspace{-3mm}
   \caption{\textbf{DINO embedding space for GLM-4~32B.} UMAP of all 1,564 valid, non-trivial generations; color is subcategory, marker shape is tier (circle T1, square T2, triangle T3). Small, tight clusters are repeated attempts at the same prompt, visually confirming the within-prompt consistency quantified numerically.}
\label{fig:glm32b_umap}
\end{figure}

If a model draws from one stable internal picture, repeated attempts at the same prompt should look similar. 
% We test this with DINO embeddings: for each prompt, we measure how tightly a model's own repeated generations cluster in embedding space relative to how far apart different prompts are.  % repetition
% GLM-4~32B, the best-performing model on the layout score, is also among the tightest by this raw measure. % we just read that
As illustrated in Figure~\ref{fig:glm32b_umap}, DINO embeddings of GLM-4~32B's generations show repeated attempts of the same prompt falling into small, well-separated clusters.

\noindent\textbf{Consistency mostly tracks sampling determinism, not spatial reliability.}
Despite the visual clustering, this raw consistency measure is heavily confounded by sampling randomness. A model sampled at temperature 0.7 with flatter next-token distributions will scatter more in any embedding space for reasons entirely unrelated to its concept representations. 
% Therefore, a raw consistency number cannot on its own distinguish genuine spatial reliability from mere output determinism. % repetition

\begin{figure}[h]
\begin{center}
\includegraphics[width=0.95\linewidth]{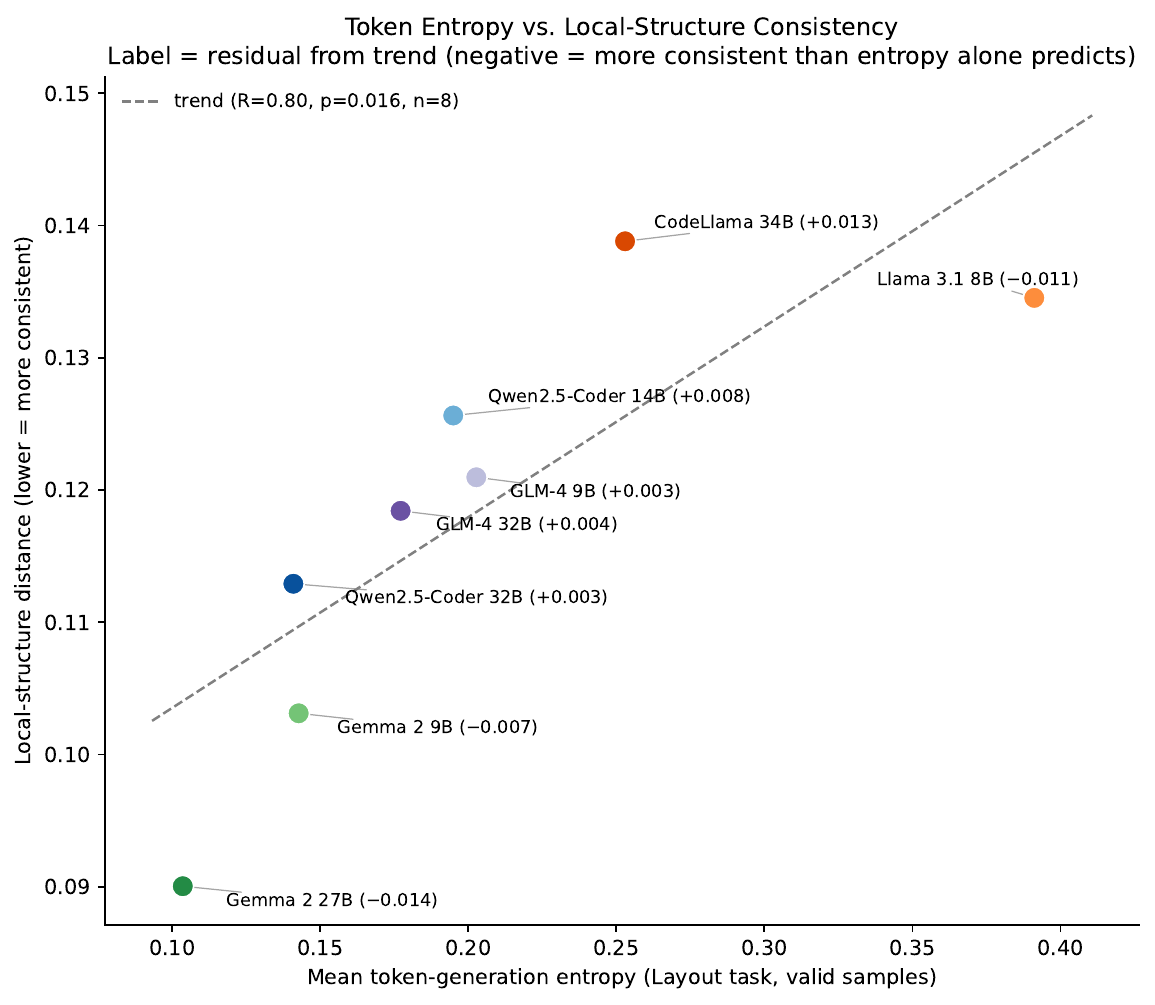}
\end{center}
%\vspace{-3mm}
   \caption{\textbf{Token-generation entropy vs.\ local-structure distance, all eight models.} Spearman $\rho=0.93$ ($p<0.001$) for rank correlation, with a linear fit of $R=0.80$ ($p=0.016$) displayed in the chart title. Point label is the residual from the fitted trend (negative $=$ more consistent than entropy alone predicts). Qwen2.5-Coder-14B's entropy values are teacher-forced; computed the same way for every other model from live generation logging.}
\label{fig:dino_entropy}
\end{figure}

We test this directly rather than by proxy: mean per-generation token entropy correlates strongly with the \emph{local}-structure distance across the eight models (Spearman $\rho{=}0.93$, $p{<}0.001$; Fig.~\ref{fig:dino_entropy}). This demonstrates that most of the cross-model spread in apparent local consistency is explained simply by how deterministic each model's sampling is.
%, meaning the raw measure alone cannot accurately rank models by spatial reliability. % does not add anything

\begin{figure}[h]
\begin{center}
\includegraphics[width=0.95\linewidth]{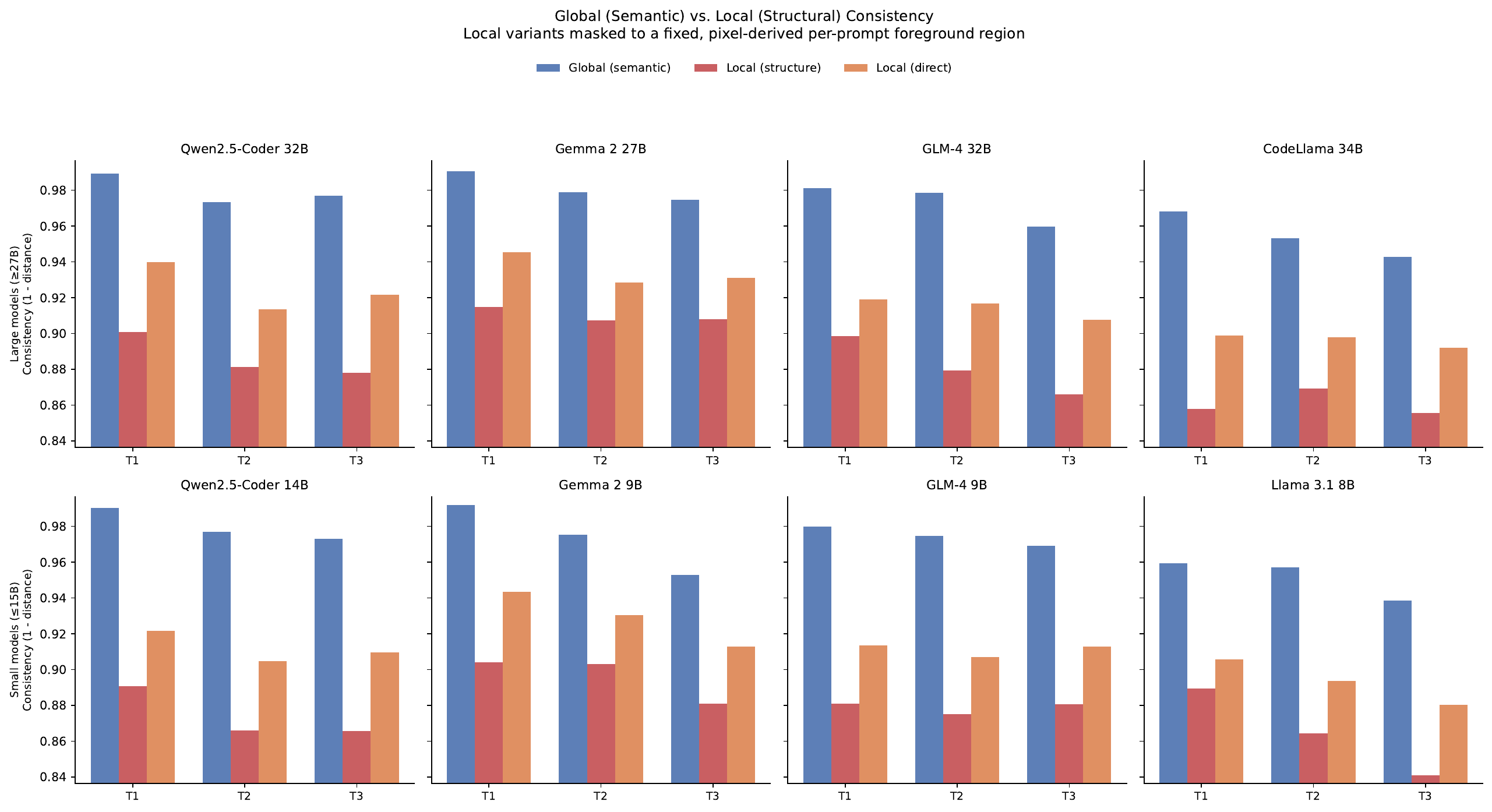}
\end{center}
%\vspace{-3mm}
   \caption{\textbf{Global (semantic) vs.\ local (structural) consistency, by model $\times$ tier, all eight models.} Higher $=$ more consistent for both measures. Divergence between the two bars for a given model $\times$ tier shows where whole-image similarity and spatial-arrangement similarity disagree.}
\label{fig:dino_global_local}
\end{figure}

\begin{figure}[h]
\begin{center}
\includegraphics[width=0.95\linewidth]{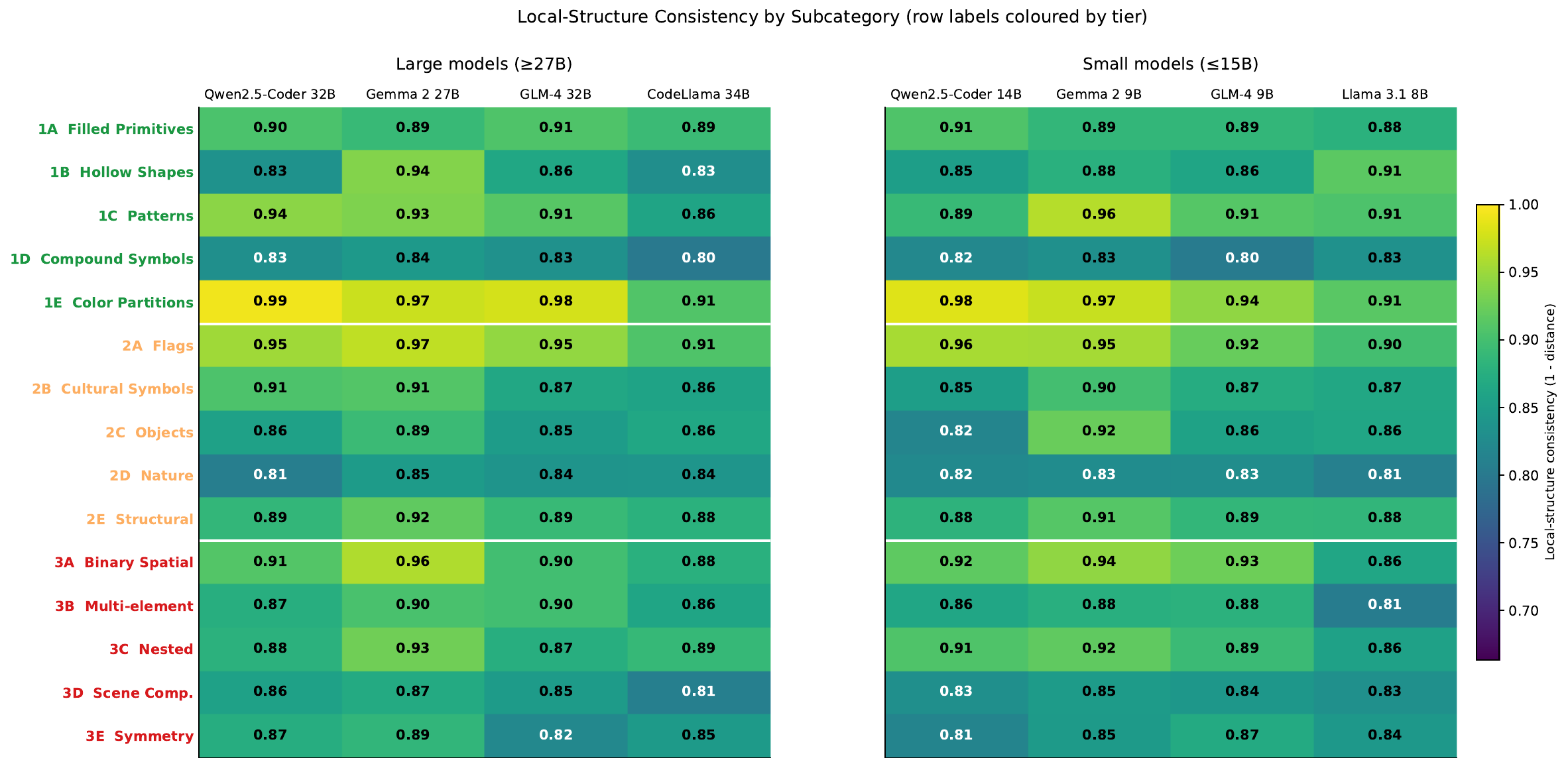}
\end{center}
%\vspace{-3mm}
   \caption{\textbf{Local-structure consistency by subcategory, all eight models.} Same tier-colored row-label convention as Fig.~\ref{fig:radar_overlay}.}
\label{fig:dino_structure_subcat}
\end{figure}

A small residual survives regressing structure distance on entropy: Gemma~2~27B is more locally consistent than its entropy alone predicts (residual $-$0.014, the largest of any model), while CodeLlama~34B is less consistent than predicted (residual $+$0.013). While this residual is real, it is several times smaller than the raw cross-model range and is too weak on its own to explain the layout-score gaps in Table~\ref{tab:scores}. 
The global-vs-local distinction itself, along with its breakdown by subcategory, is detailed below (Fig.~\ref{fig:dino_global_local}, Fig.~\ref{fig:dino_structure_subcat}).

\begin{figure*}[t]
\begin{center}
\includegraphics[width=0.8\linewidth]{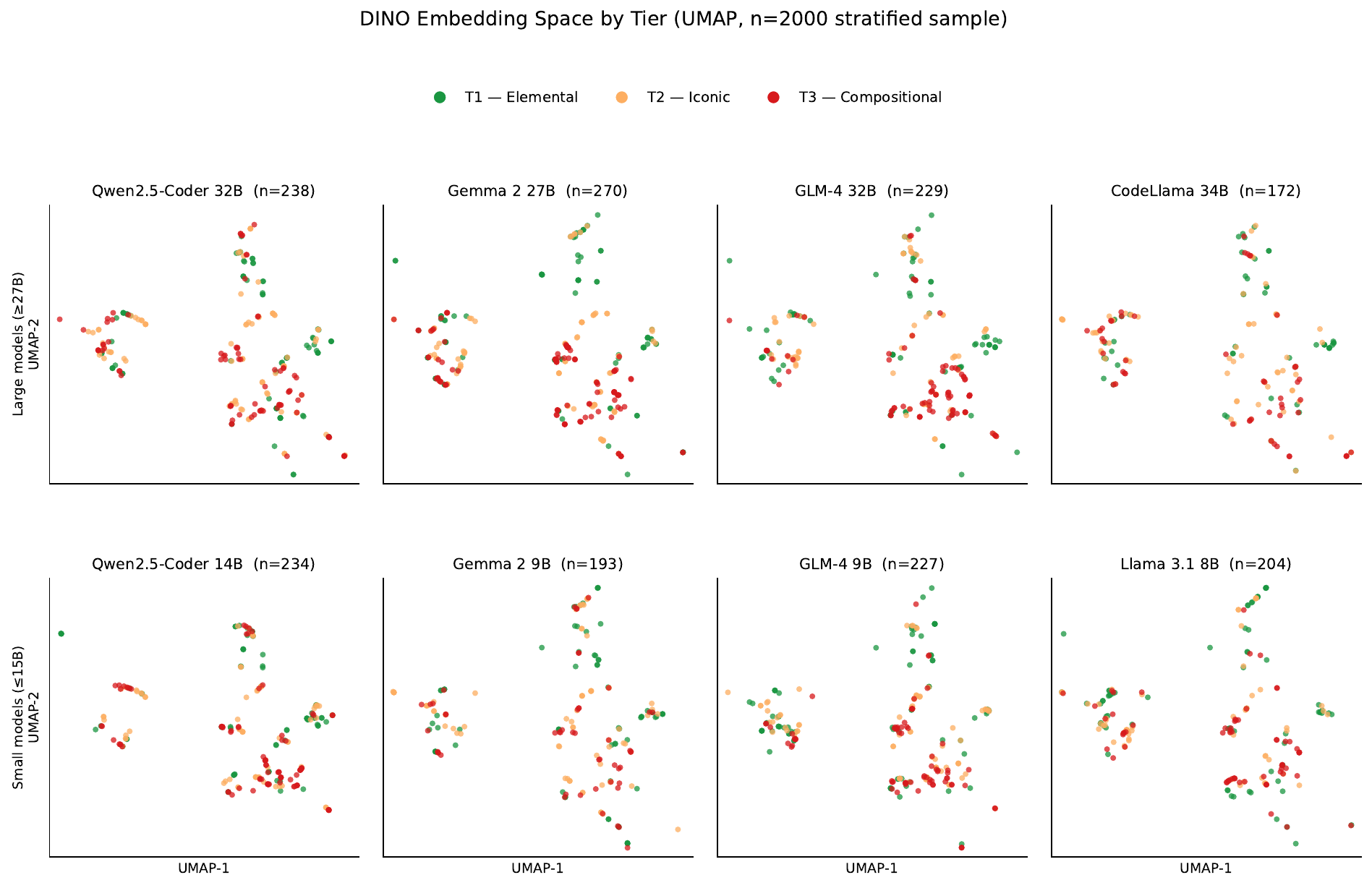}
\end{center}
%\vspace{-3mm}
   \caption{\textbf{UMAP of DINO embeddings, all eight models.} $\approx$2,000-image stratified subsample, colored by tier only. T1 separates cleanly; T2/T3 overlap more with each other than with T1.}
\label{fig:dino_umap}
\end{figure*}

\noindent\textbf{UMAP embedding space.} Fig.~\ref{fig:dino_umap} shows tier structure largely preserved in DINO space on a $\approx$2,000-image stratified subsample: T1 embeddings separate cleanly from T2/T3, so visual complexity, more than model identity, is the dominant axis DINO captures. T2 and T3 overlap more with each other than either does with T1, consistent with Table~\ref{tab:dino_coherence}'s per-tier coherence numbers.

\noindent\textbf{Reading this analysis with the translation gate.} Translation is not the bottleneck for any model (Appendix~\ref{app:translation_details}). Exp.~2 gives a second, independent signal, with two caveats now made explicit that the raw global measure alone would have missed: the two most globally scattered models, Llama~3.1~8B and CodeLlama~34B, are also the two lowest-ranked on layout score (Table~\ref{tab:scores}), consistent with unstable composition rather than a translation problem, but a meaningful share of that raw scatter is explained by how deterministic each model's sampling is (above), not by spatial reliability alone. A direct answer at the representation level, before generation happens at all, is what Exp.~3 tests (Sec.~\ref{sec:exp3}). An illustrative, not quantified, view of attention concentration is in Appendix~\ref{app:dino_attention}.

\noindent\textbf{Global vs.\ local structure, full detail.}
The local-structure metric restricts DINO patch-key self-similarity to a per-prompt foreground mask, pooled across generations at a majority-vote threshold (Appendix~\ref{app:dino}), rather than the global metric's whole-image CLS embedding, so it is sensitive specifically to spatial arrangement rather than to overall semantic content (object identity, color, composition). 

\subsection{DINO attention maps (illustrative)}
\label{app:dino_attention}
\begin{figure*}[hbt!]
\begin{center}
\includegraphics[width=0.8\linewidth]{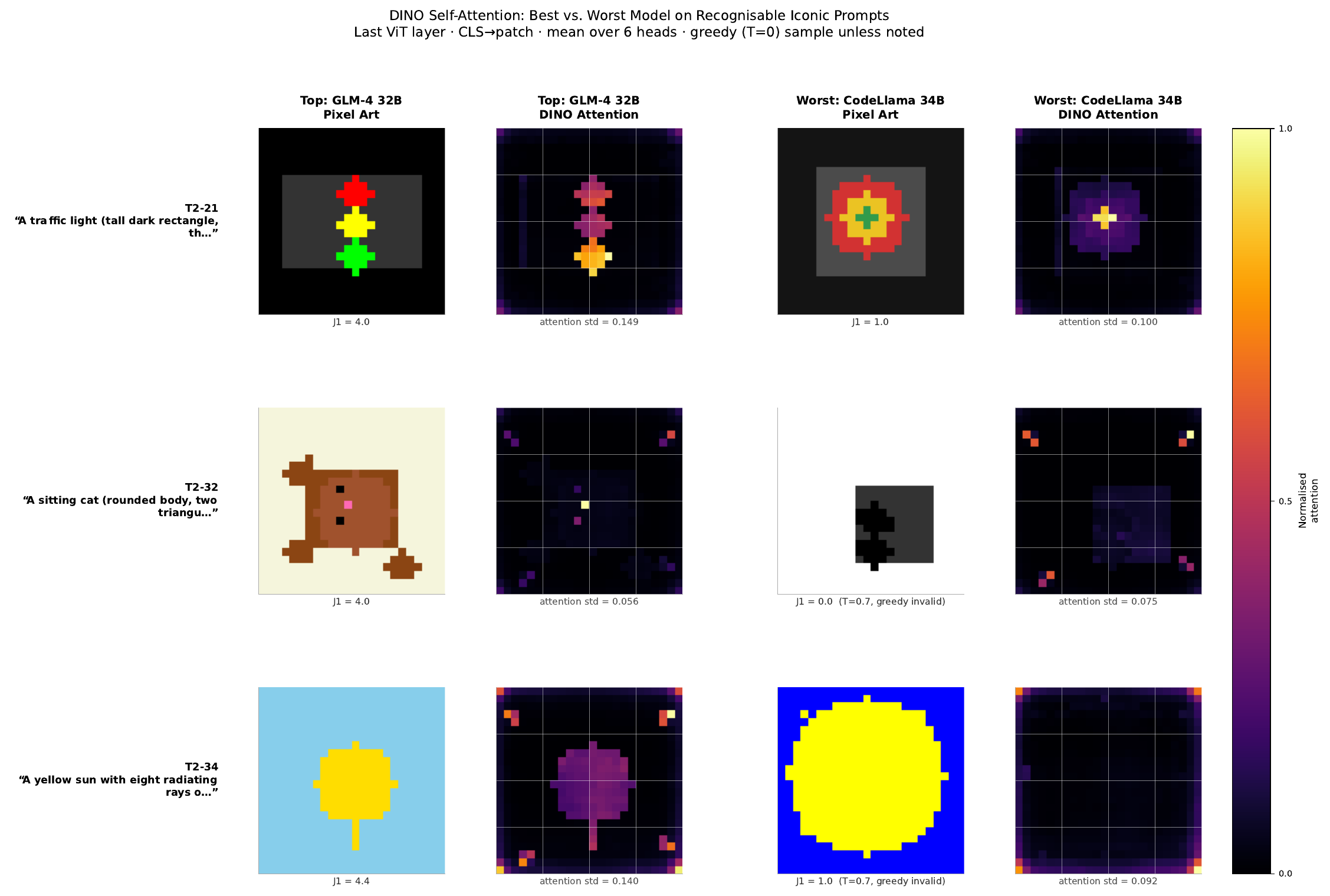}
\end{center}
%\vspace{-3mm}
   \caption{\textbf{DINO attention maps, illustrative (see Appendix~\ref{app:dino_attention}).} Top model (GLM-4~32B) vs.\ bottom model (CodeLlama~34B) by layout score, three Iconic-tier prompts each, hand-picked most-focused example; not a quantified average.}
\label{fig:dino_attention}
\end{figure*}

Fig.~\ref{fig:dino_attention} visualizes DINO last-block CLS-to-patch attention for the top- and bottom-ranked models by mean layout score (GLM-4~32B and CodeLlama~34B respectively, Table~\ref{tab:scores}) on the same Iconic-tier prompts, one selected generation per model chosen as the highest attention-map standard deviation from a pool of candidates. This selection procedure picks the most spatially-focused example available and is illustrative of what focused versus diffuse attention looks like; it is not a quantified claim about typical behavior across all generations, since selecting a maximum and describing it as representative would be circular. The pattern shown, a tighter focus for GLM-4~32B, more diffuse attention for CodeLlama~34B, is directionally consistent with the quantitative coherence collapse in Table~\ref{tab:dino_coherence} (CodeLlama~34B is one of only two models with negative silhouette at every tier), but should be read as a visual illustration of that finding, not as independent evidence for it.

\section{Metadata \& Reproducibility}
\label{app:meta}

\subsection{Statistical plan and reproducibility}
\label{app:stats}
VLM scores are compared by tier and model with the summary statistics reported in the main text (mean, median, per-dimension breakdown); primary comparisons (translation task pass/fail, Exp.~3's eight per-model decodability gaps, and the causal arm's per-metric contrasts) use resampling rather than a parametric test, reported with explicit confidence intervals rather than $p$-values alone where a CI is the more informative object, and with Benjamini-Hochberg correction wherever more than one cell is tested. Every seeded operation in the translation stress-test, the null baseline, and the Exp.~3 pipelines takes an explicit integer seed passed as an argument; no operation relies on Python's default-randomised \texttt{hash()} or an unseeded random-number generator, a discipline adopted after an early version of the null-baseline code used \texttt{hash()} for a per-reference seed and was found, on rerunning, to give slightly different numbers each time. Repeated runs of the translation stress-test, the null baseline, and the full Exp.~3 sweep reproduce the reported numbers exactly on the same hardware; the MLP-probe path used in the superseded first-pass analysis (unlike the linear-probe path) is not guaranteed bitwise-identical across CPU and GPU, since the two execute floating-point matrix operations via different underlying kernels even with every seed fixed. This affects point estimates by well under 0.01 in the cases checked, not any significance conclusion.

\subsection{Limitations, full detail}
\label{app:limitations_full}
Full detail behind the summary in Sec.~\ref{sec:limitations}.

\textbf{Four families, two size points each.} The roster spans Qwen, Gemma, GLM, and Llama/CodeLlama, 8B--34B, which is far broader than a single-family pilot, but it is still not a controlled scaling-law sweep: each family contributes exactly two sizes, chosen for availability rather than even log-spacing, and four of the eight (Gemma~2~27B/9B, CodeLlama~34B, Llama~3.1~8B) are gated weights whose inclusion depended on license acceptance rather than a fixed sampling frame. Whether the patterns here (translation solved cleanly, an Iconic/Compositional inversion, a mostly-null decodability gap over text) hold at larger scale within a family, or for closed-weight models we could not run at all, is untested.

\textbf{The gate's scope.} The translation task rules out code-writing competence as an explanation for a low layout score; it does not by itself separate an absent internal layout from one present but not convertible into primitives (Sec.~\ref{sec:layout_design}). Exp.~3 addresses this directly and finds a coarse layout decodable above a text baseline for all eight models (Sec.~\ref{sec:exp3}). What is decodable is the layout the prompt implies; the model-specific realisation is not, against a measured ceiling of 0.88--0.94. The evidence therefore points toward a generic plan that is present and a specific one that is settled during generation, rather than a specific plan that is present but unconvertible.

\textbf{Numeric tokenization is a candidate confound the translation task does not directly rule out.} Standard subword tokenizers represent numbers inconsistently; continuous alternatives have been proposed specifically to address this for scientific and numeric tasks \cite{golkar2023xval}. Since every primitive call here takes explicit numeric coordinates, a model whose tokenizer handles numbers poorly could, in principle, under-perform at Layout for reasons unrelated to spatial composition. The translation task offers indirect evidence against this being severe for the eight models studied here: transcribing a stated geometry into primitives requires writing the same kind of precise numeric coordinates, and the median PIoU is 1.0 for every model (Table~\ref{tab:translation_headline}), so gross numeric-tokenization failure is not apparent in this roster. A dedicated ablation (e.g.\ coordinates in alternative encodings) would be needed to rule it out directly.

\textbf{The ladder tests geometric difficulty, not periodicity, and is two-model-only.} The 100-reference stress-test ladder and 13-icon set (Appendix~\ref{app:translation_details}) were built and run for the original two-model pilot, Qwen2.5-Coder-14B/7B, before the roster grew to eight, and were not re-run for the other six models; the 145-reference task result in Table~\ref{tab:translation_headline} is the only translation result that covers all eight. Within the pilot pair, the ladder's scores run slightly \emph{higher} than the first-round set, not lower; its four axes are all axes of geometric difficulty, and periodicity, repeating grids or tilings, was never built into it. Given 7B's weakest Layout subcategories are Patterns and Symmetry, periodicity is the most likely axis on which translation would actually degrade, and it remains untested for any model.

\textbf{Reference quality.} 17 of the 145 translation targets needed simplification (merging touching, same-color shapes) before they could be described unambiguously; within the two-model pilot, both models score measurably worse on these (0.96 vs.\ 1.00 for 14B; 0.90 vs.\ 0.98 for 7B). A modest, documented source of noise in the target set, not enough to change the outcome for any of the eight models (Table~\ref{tab:translation_headline}).

\textbf{Three data-quality notes from a full sanity sweep of every result reported here}, kept brief; (i)~Qwen2.5-Coder-14B's Layout-task generation-entropy values (used in Fig.~\ref{fig:dino_entropy}) were computed by teacher-forcing over its already-saved code rather than live logging, since that model's Layout data is the original pilot run and predates entropy instrumentation; mathematically equivalent to a live measurement, but not an identical reproduction if the model's original raw completion included wrapper text later stripped by parsing. (ii)~Six of 12,932 valid Layout samples (0.05\%) are missing a second-judge score; five cluster on a single prompt (a clock face) across four different models, plausibly a detail-density effect on the judge's output parsing rather than a systemic issue; every reported mean skips these automatically. (iii)~The pooled foreground mask behind the local-structure metric (Appendix~\ref{app:dino_consistency}) is empty for $\approx$3\% of structure-coherence cells, concentrated on prompts describing several small, spatially diffuse elements (e.g.\ ``seven small squares arranged in a ring''), where no single pixel is foreground in enough pooled generations to clear the 50\% threshold, a real blind spot of majority-vote pooling rather than corrupted data.

\textbf{Scope.} AM-Bench measures allocentric 2D layout at mosaic granularity. It is a necessary-condition probe for embodied spatial competence, not a sufficient one, and it does not cover metric egocentric continuity, depth, or occlusion.

\textbf{Future work.} Several avenues remain for expanding this work. First, the benchmark itself can be scaled to encompass broader spatial categories, supported by a larger multi-annotator human validation run and the extension of both the translation stress-test ladder (including a novel periodicity rung) and the icon set across the full eight-model roster. To further investigate the effect of the output medium (Sec.~\ref{sec:exp_medium}), future iterations could introduce other medium designs as an additional code-prior ablation. This will be paired with rigorous memorization controls, utilizing held-out rare compositions and a matched-scale distillation baseline. Finally, to substantiate Exp.~3's behavioral claims at the representation level, we intend to apply activation patching for a direct, causal analysis of the models' internal spatial planning.

\subsection{Code Release}
\label{app:release}
All prompts, along with the code (executor, evaluation pipeline, translation scorer, DINO analysis, experiments) for generating the dataset and reproducing the experiments, will be publicly released. See the \textbf{Project Page} for more.
\end{document}